\documentclass{article} %
\usepackage{arxiv}

\usepackage{amsmath,amsfonts,bm}

\def\eqref#1{equation~\ref{#1}}

\def\1{\bm{1}}

\DeclareMathAlphabet{\mathsfit}{\encodingdefault}{\sfdefault}{m}{sl}
\SetMathAlphabet{\mathsfit}{bold}{\encodingdefault}{\sfdefault}{bx}{n}

\usepackage{hyperref}
\usepackage{url}
\usepackage{graphicx} 
\usepackage{subcaption}
\usepackage{amssymb}
\usepackage{booktabs}
\usepackage{multirow}
\usepackage[table]{xcolor}
\definecolor{effectbg}{RGB}{242,244,247}
\definecolor{benchmarkbg}{RGB}{214,226,241}
\definecolor{baselinebg}{RGB}{245,245,245}
\definecolor{baselineframe}{RGB}{180,180,180}
\definecolor{accentred}{RGB}{166,57,48}
\usepackage{array}
\usepackage{threeparttable}
\usepackage[table]{xcolor}

\usepackage{enumitem}
\usepackage[most]{tcolorbox}

\definecolor{effectbg}{RGB}{246,247,249}
\newcolumntype{d}{>{\columncolor{effectbg}}c}

\title{Formatting Instructions for ICLR 2027 \\ Conference Submissions}

\author{Minghan Wang\textsuperscript{$\diamond, \ast$}, Boyuan Wang\textsuperscript{$\diamond, \ast$}, Jinhang Zuo\textsuperscript{$\circ$}, Yuxin Tao\textsuperscript{$\diamond, \dagger$}, Fang Kong\textsuperscript{$\diamond, \dagger$}}

\title{
Thinking Outside the Box: Can Language Models Rely on External Guidance Selectively?
}

\newtcolorbox{prompt}[1]{
  enhanced,
  breakable,
  colback=baselinebg,
  colframe=baselineframe,
  boxrule=0.5pt,
  arc=1.2mm,
  title={#1},
  fonttitle=\bfseries,
  before skip=0.8em,
  after skip=0.8em,
  left=1.2em,
  right=1.2em,
  top=0.8em,
  bottom=0.8em
}

\begin{document}

\maketitle

\begin{center}
    \vspace{-2mm}
    \textsuperscript{$\diamond$} Southern University of Science and Technology
    ,
    \textsuperscript{$\circ$} City University of Hong Kong
\end{center}

\begin{abstract}
Agent harnesses often improve language models with human-designed workflows, but as models grow more capable, unreliable guidance can increasingly constrain their execution.
We call the ability to benefit from useful guidance while overriding unreliable guidance \textbf{thinking outside the
box}.
We introduce Box$^2$-Bench, which holds the model and task fixed while varying workflow reliability to isolate how models regulate their reliance on guidance. 
On Box$^2$-Bench, frontier models often benefit from reliable guidance but remain vulnerable when it is misleading or becomes unreliable.
To test whether this capability can be learned, we train two open-weight models using bad workflows, reserving good workflows for evaluation. 
We explore two complementary training strategies: counterfactual supervised fine-tuning improves robustness, while outcome-based reinforcement learning can shift the balance toward greater use of helpful workflows.
We further find that this behavior extends beyond workflows to other forms of external information, improving peer correction and robustness to corrupted memory. Together, our results identify selective reliance on fallible external information as a dimension of agent reliability not captured by task performance alone.
\end{abstract}

\begingroup
\renewcommand{\thefootnote}{\fnsymbol{footnote}}

\footnotetext[1]{Equal contribution; order decided by a coin flip.}
\footnotetext[2]{Correspondence to 
\href{taoyx@sustech.edu.cn}{taoyx@sustech.edu.cn} and
\href{kongf@sustech.edu.cn}{kongf@sustech.edu.cn}.}
\endgroup

\begin{figure*}[h]
\centering

\includegraphics[width=0.97\textwidth]{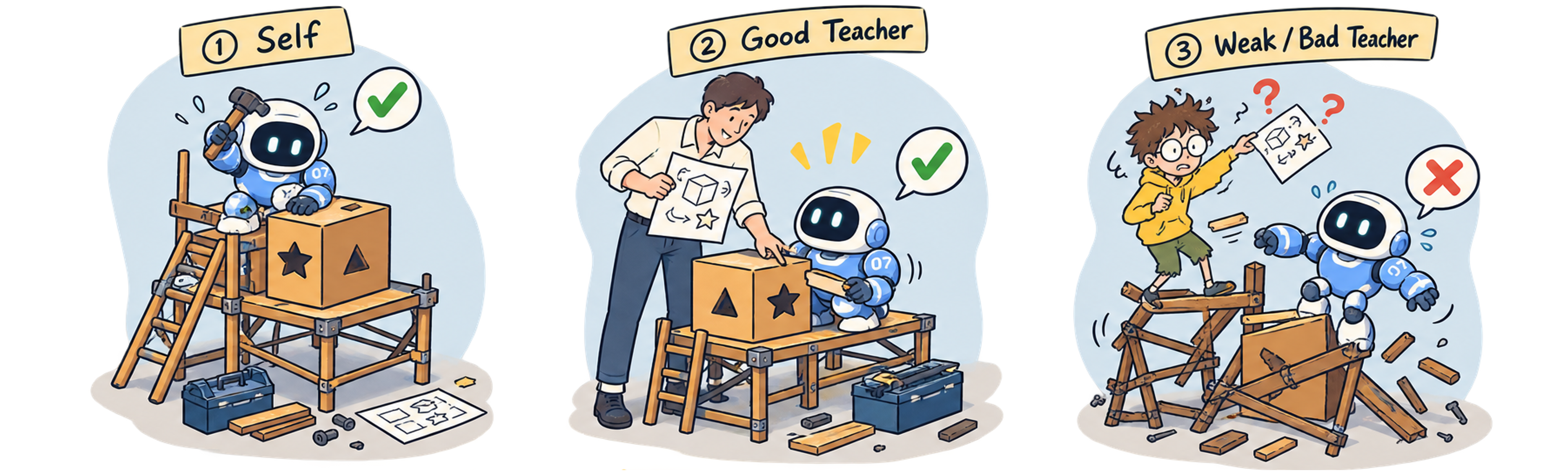}

\vspace{0.5em}

\begin{subfigure}[t]{0.45\textwidth}
\centering
\includegraphics[width=\linewidth]{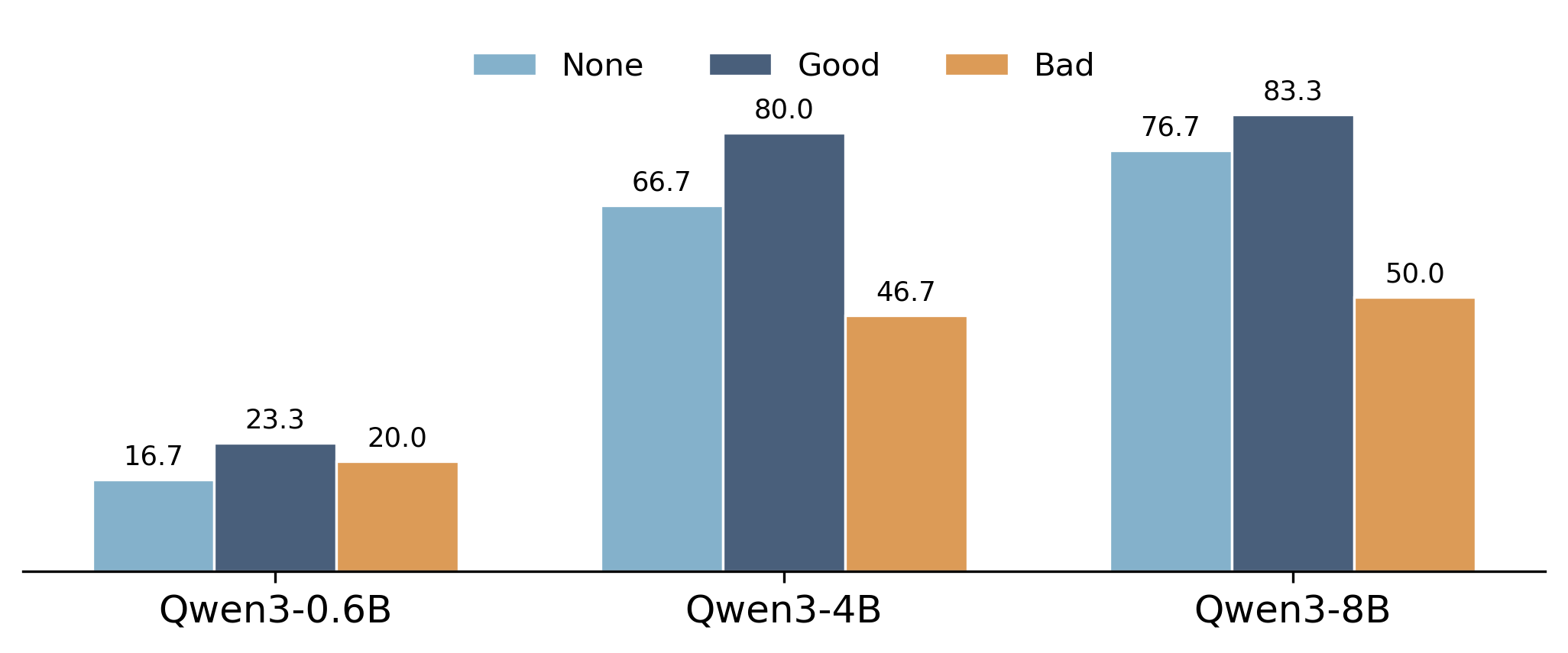}
\caption{AIME 2026}
\label{fig:aime-harness}
\end{subfigure}
\hspace{0.01\textwidth}
\begin{subfigure}[t]{0.45\textwidth}
\centering
\includegraphics[width=\linewidth]{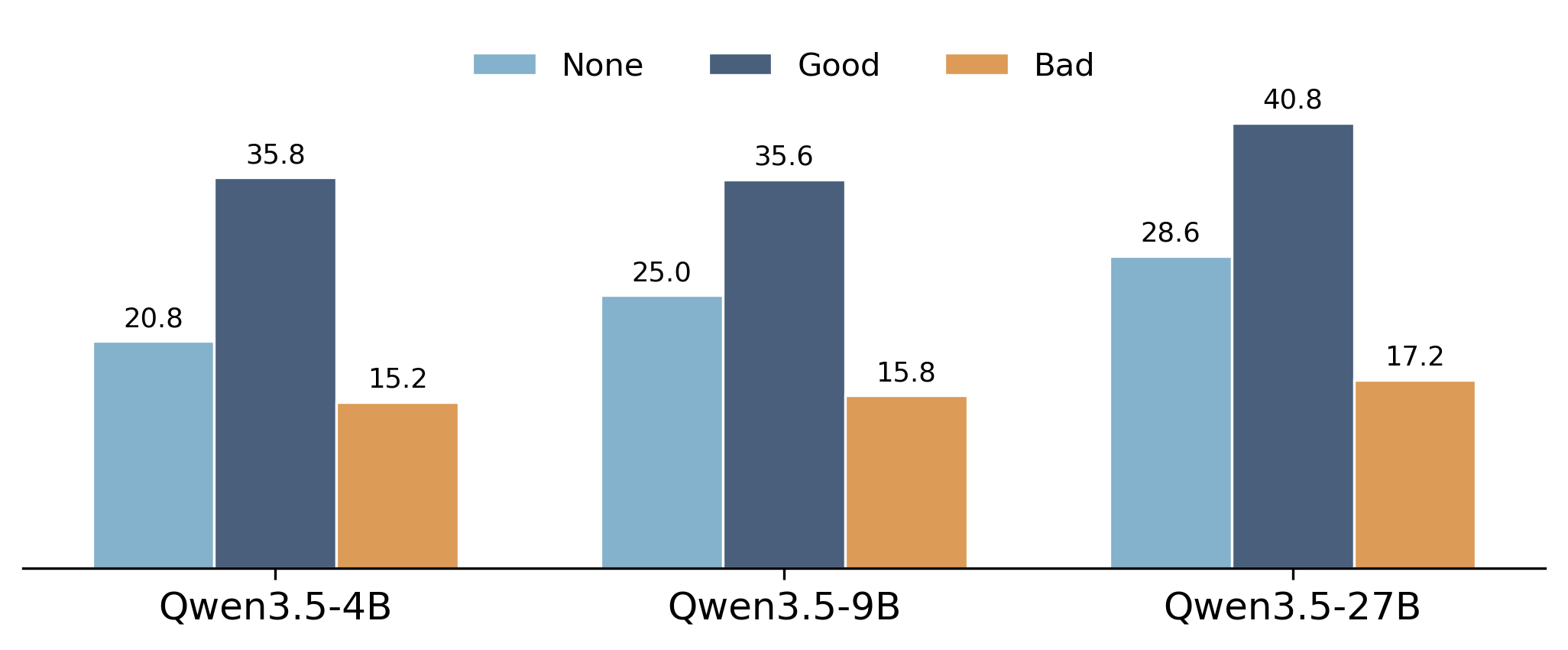}
\caption{WebShop}
\label{fig:webshop-harness}
\end{subfigure}

\caption{
The top illustration contrasts independent execution with guidance from capable and weak teachers.
The bottom panels show performance under self-solving, good-workflow, and bad-workflow conditions on
\textbf{(a)} AIME 2026 and \textbf{(b)} WebShop.
Good workflows improve performance, while bad ones expose sensitivity to misleading guidance.
}
\label{fig:harness-motivation}

\end{figure*}

\section{Introduction}

Large language models increasingly operate within harnesses that structure planning, feedback, and tool use
\citep{li2026flow,ruan2026aorchestra,zhang2025aflow,shang2025agentsquare}.
Harnesses inject procedural priors, guiding models that cannot yet discover reliable strategies on their own
\citep{sarukkai2025selfgenerated,zhou2026enhancing,wang2026harnesses}.
This benefit, however, assumes that the human guide is more reliable than the model. 
As frontier models solve increasingly complex problems in reasoning, coding, and long-horizon research
\citep{feng2026towards,kung2026leap,huang2026deepswe,yamada2026towards,ma2026lower},
this assumption becomes increasingly important to revisit.

When a model can find a better strategy on its own, following a prescribed workflow can constrain its execution. Removing procedural guidance from the harness creates the opposite problem, since useful guidance can still improve performance. \textbf{We call this ability \emph{thinking outside the box}: following external procedural guidance when it helps, resisting it when it misleads, and moving beyond it when a better strategy emerges.} This selective workflow reliance is distinct from task-solving capability: the latter asks whether a model can solve the underlying task, whereas the former asks whether it can benefit from external procedural knowledge without becoming bound by it. Figure~\ref{fig:harness-motivation} illustrates this tension.

Existing benchmarks evaluate autonomous task completion and compliance with
prescribed procedures, 
but not whether models can adjust their reliance on guidance as its reliability changes
\citep{merrill2026terminalbench,li2026agencybench,deng2025swe,
wang2026sopmaze,cao2026procedureaware,guan2026supchain}.
\textbf{We introduce Box$^2$-Bench to evaluate \emph{thinking outside the box}.}
Box$^2$ holds the task, environment, evaluator, and model fixed while varying
workflow availability and reliability across five matched conditions:
\begin{equation}
\label{eq:box2-conditions}
\begin{array}{c@{\qquad}c@{\qquad}c@{\qquad}c@{\qquad}c}
\textbf{No workflow} & \textbf{Good} & \textbf{Partial} &
\textbf{Mixed} & \textbf{Bad} \\[2pt]
\varnothing\,\varnothing\,\varnothing\,\varnothing\,\varnothing &
+\,+\,+\,+\,+ &
+\,+\,+\,\varnothing\,\varnothing &
+\,+\,+\,-\,- &
-\,-\,-\,-\,-
\end{array},
\nonumber
\end{equation}
where $+$, $-$, and $\varnothing$ denote useful, misleading, and absent
guidance, respectively. Thus, Good and Bad workflows provide consistently useful or misleading guidance, while Partial and Mixed workflows share a useful prefix but then either stop or become misleading.
A fixed external model serves as a scalable proxy for human workflow designers,
and each generated workflow is independently verified.
On Box$^2$-Bench, across frontier models and diverse tasks, we find a consistent capability gap:
models benefit from useful workflows yet remain vulnerable when similar
guidance is misleading or becomes unreliable. Task performance alone therefore
does not capture a model's ability to regulate its reliance on external
guidance.

We next ask whether models can learn to reject bad guidance without rejecting guidance altogether.
We train open-weight models on bad workflows, reserving good workflows for evaluation. 
Counterfactual SFT teaches models to override guidance that conflicts with task success, while outcome-based RL rewards success without prescribing a trajectory. 
SFT improves robustness but induces overly broad workflow resistance; RL partially restores use of held-out good workflows while retaining a robustness improvement over the base model. 
\textbf{These results show that robustness to bad guidance is not enough: thinking outside the box requires selective reliance rather than blanket rejection.}

We further ask whether this selectivity is specific to workflows or extends to other forms of fallible information.
Without setting-specific training, we evaluate the same models with peer information in multi-agent reasoning and stored information in memory-augmented reasoning.
In multi-agent collaboration, training leads to more beneficial cross-agent corrections. 
In memory-augmented reasoning, models preserve the benefits of reliable memory while more often verifying corrupted records. 
Together, these results provide initial evidence that learning to regulate workflow reliance can generalize to how models use external information more broadly.

\begin{table*}[t]
\centering
\caption{
Box$^2$ performance across five execution regimes.
We first report absolute performance under each regime, followed by
the corresponding paired workflow effects.
}
\label{tab:box2-main-frontier}

\small
\setlength{\tabcolsep}{6.5pt}
\renewcommand{\arraystretch}{1.05}
\resizebox{\linewidth}{!}{
\begin{tabular}{lcccccdddd}
\toprule

\multirow{3}{*}{\textbf{Model}}
&
\multicolumn{5}{c}{\textbf{Performance}}
&
\multicolumn{4}{c}{\textbf{Box$^2$-Bench Effects}}
\\

\cmidrule(lr){2-6}
\cmidrule(lr){7-10}

&
\textit{Base}
&
\textit{Good}
&
\textit{Bad}
&
\textit{Partial}
&
\textit{Mixed}
&
\cellcolor{effectbg}\textit{Utilization}
&
\cellcolor{effectbg}\textit{Robustness}
&
\multicolumn{2}{c}{\cellcolor{effectbg}\textit{Recovery}}
\\

&
$S_0$
&
$S_G$
&
$S_B$
&
$S_P$
&
$S_M$
&
\cellcolor{effectbg}$\Delta_{\mathrm{use}}$
&
\cellcolor{effectbg}$\Delta_{\mathrm{bad}}$
&
\cellcolor{effectbg}$\Delta_{\mathrm{stop}}$
&
\cellcolor{effectbg}$\Delta_{\mathrm{switch}}$
\\[-1pt]

\midrule

\rowcolor{benchmarkbg}
\multicolumn{10}{l}{
    \rule{0pt}{2ex}
    \small \# \textit{Math --- OpenR1-Math}
}\\

\textbf{Gemini 3.7 Flash}
    & 76.7\% & 83.3\% & 33.3\% & 80.0\% & 43.3\%
    & +6.7 & -43.3 & -3.3 & -36.7 \\

\textbf{DeepSeek V4 Flash}
    & 76.7\% & 80.0\% & 66.7\% & 83.3\% & 73.3\%
    & +3.3 & -10.0 & +3.3 & -10.0 \\

\textbf{GLM 5.2}
    & 80.0\% & 80.0\% & 40.0\% & 83.3\% & 56.7\%
    & 0.0 & -40.0 & +3.3 & -26.7 \\

\midrule[0.35pt]

\rowcolor{benchmarkbg}
\multicolumn{10}{l}{
    \rule{0pt}{2ex}
    \small \# \textit{Code --- DeepSWE}
}\\

\textbf{Gemini 3.7 Flash}
    & 60.0\% & 53.3\% & 50.0\% & 60.0\% & 56.7\%
    & -6.7 & -10.0 & +6.7 & -3.3 \\

\textbf{DeepSeek V4 Flash}
    & 26.7\% & 36.7\% & 13.3\% & 36.7\% & 26.7\%
    & +10.0 & -13.3 & 0.0 & -10.0 \\

\textbf{GLM 5.2}
    & 23.3\% & 23.3\% & 20.0\% & 16.7\% & 23.3\%
    & 0.0 & -3.3 & -6.7 & +6.7 \\

\midrule[0.35pt]

\rowcolor{benchmarkbg}
\multicolumn{10}{l}{
    \rule{0pt}{2ex}
    \small \# \textit{Information Search --- BrowseComp}
}\\

\textbf{Gemini 3.7 Flash}
    & 46.7\% & 46.7\% & 36.7\% & 36.7\% & 36.7\%
    & 0.0 & -10.0 & -10.0 & 0.0 \\

\textbf{DeepSeek V4 Flash}
    & 20.0\% & 23.3\% & 10.0\% & 23.3\% & 16.7\%
    & +3.3 & -10.0 & 0.0 & -6.7 \\

\textbf{GLM 5.2}
    & 13.3\% & 16.7\% & 10.0\% & 16.7\% & 13.3\%
    & +3.3 & -3.3 & 0.0 & -3.3 \\

\midrule[0.35pt]

\rowcolor{benchmarkbg}
\multicolumn{10}{l}{
    \rule{0pt}{2ex}
    \small \# \textit{Tool Use --- AutomationBench}
}\\

\textbf{Gemini 3.7 Flash}
    & 74.4\% & 80.3\% & 65.6\% & 84.8\% & 77.4\%
    & +5.9 & -8.8 & +4.5 & -7.3 \\

\textbf{DeepSeek V4 Flash}
    & 69.3\% & 83.0\% & 57.3\% & 83.6\% & 70.4\%
    & +13.7 & -12.0 & +0.6 & -13.2 \\

\textbf{GLM 5.2}
    & 45.6\% & 62.4\% & 43.2\% & 62.9\% & 54.2\%
    & +16.8 & -2.3 & +0.5 & -8.7 \\

\bottomrule
\end{tabular}
}
\begin{tablenotes}[flushleft]
\footnotesize
\item[]
\textit{Note.}
Effects measure good-guidance utilization
($\Delta_{\mathrm{use}}=S_G-S_0$),
bad-guidance robustness
($\Delta_{\mathrm{bad}}=S_B-S_0$),
and recovery when guidance stops
($\Delta_{\mathrm{stop}}=S_P-S_G$)
or becomes misleading
($\Delta_{\mathrm{switch}}=S_M-S_P$).
\end{tablenotes}
\end{table*}

\section{Box$^2$-Bench: Benchmarking Inside- and Outside-the-Box Execution}
\label{sec:box2-revised}

Box$^2$-Bench evaluates whether models benefit from useful workflow guidance while remaining robust to misleading guidance. 
Raw task accuracy does not reveal this distinction. 
We therefore hold the model and task fixed and vary only the supplied workflow, isolating the effect of workflow reliability from task difficulty.

\subsection{Benchmark Design}
\label{sec:box2-revised-setup}

\textbf{Intuition.}
We view workflow-conditioned task solving at two levels. The workflow provides
an outer sequence of goals, while the model controls the inner trajectory used
to complete each goal. Thinking outside the box requires following this outer
plan when it is reliable and departing from it when it becomes incomplete or
misleading.

\textbf{Workflow-conditioned execution.}
We follow the two-level formulation of harnessed execution in
\citet{wang2026harnesses} and focus on its workflow component. For a task
$x\sim\mathcal{D}$, we represent a workflow and its resulting execution as
\begin{equation}
W(x)=(g_1,\ldots,g_T),
\qquad
\tau_W(x)=(g_1,\tau_1,\ldots,g_T,\tau_T),
\end{equation}
where $\tau_t$ is the model trajectory associated with goal $g_t$. Box$^2$
holds the task and model fixed while intervening only on the workflow $W(x)$.

\begin{table*}[t]
\centering
\caption{
\textbf{Box$^2$-Bench performance across model scales on AIME 2026 and WebShop.}
Good workflows consistently improve performance, whereas bad workflows generally reduce it.
Scaling does not reliably eliminate sensitivity to misleading guidance.
}
\label{tab:box2-main}

\small
\setlength{\tabcolsep}{6.5pt}
\renewcommand{\arraystretch}{1.05}
\resizebox{\linewidth}{!}{
\begin{tabular}{lcccccdddd}
\toprule

\multirow{3}{*}{\textbf{Model}}
&
\multicolumn{5}{c}{\textbf{Performance}}
&
\multicolumn{4}{c}{\textbf{Box$^2$-Bench Effects}}
\\

\cmidrule(lr){2-6}
\cmidrule(lr){7-10}

&
\textit{Base}
&
\textit{Good}
&
\textit{Bad}
&
\textit{Partial}
&
\textit{Mixed}
&
\cellcolor{effectbg}\textit{Utilization}
&
\cellcolor{effectbg}\textit{Robustness}
&
\multicolumn{2}{c}{\cellcolor{effectbg}\textit{Recovery}}
\\

&
$S_0$
&
$S_G$
&
$S_B$
&
$S_P$
&
$S_M$
&
\cellcolor{effectbg}
$\Delta_{\mathrm{use}}$
&
\cellcolor{effectbg}
$\Delta_{\mathrm{bad}}$
&
\cellcolor{effectbg}
$\Delta_{\mathrm{stop}}$
&
\cellcolor{effectbg}
$\Delta_{\mathrm{switch}}$
\\[-1pt]

\midrule

\rowcolor{benchmarkbg}
\multicolumn{10}{l}{
    \rule{0pt}{2ex}
    \small \# \textit{Math --- AIME2026}
}\\

\textbf{Qwen3-0.6B}
    & 16.7\%
    & 23.3\% & 20.0\% & 15.6\% & 15.6\%
    & +6.7 & +3.3 & -7.8 & +0.0 \\

\textbf{Qwen3-4B}
    & 66.7\%
    & 80.0\% & 46.7\% & 81.1\% & 66.7\%
    & +13.3 & -20.0 & +1.1 & -14.4 \\

\textbf{Qwen3-8B}
    & 76.7\%
    & 83.3\% & 50.0\% & 84.4\% & 74.4\%
    & +6.7 & -26.7 & +1.1 & -10.0 \\

\midrule[0.35pt]

\rowcolor{benchmarkbg}
\multicolumn{10}{l}{
    \rule{0pt}{2ex}
    \small \# \textit{Search --- WebShop}
}\\

\textbf{Qwen3.5-4B}
    & 20.8\%
    & 35.8\%
    & 15.2\%
    & 33.3\%
    & 30.2\%
    & +15.0
    & -5.6
    & -2.5
    & -3.1 \\

\textbf{Qwen3.5-9B}
    & 25.2\%
    & 35.6\%
    & 15.4\%
    & 32.7\%
    & 32.6\%
    & +10.4
    & -9.8
    & -2.9
    & -0.1 \\

\textbf{Qwen3.5-27B}
    & 28.6\%
    & 40.8\%
    & 17.2\%
    & 39.0\%
    & 39.5\%
    & +12.2
    & -11.4
    & -1.8
    & +0.5 \\

\bottomrule
\end{tabular}
}
\begin{tablenotes}[flushleft]
\footnotesize
\item[]
\textit{Note.}
Effects measure good-guidance utilization
($\Delta_{\mathrm{use}}=S_G-S_0$),
bad-guidance robustness
($\Delta_{\mathrm{bad}}=S_B-S_0$),
and recovery when guidance stops
($\Delta_{\mathrm{stop}}=S_P-S_G$)
or becomes misleading
($\Delta_{\mathrm{switch}}=S_M-S_P$).
\end{tablenotes}
\end{table*}

\begin{figure}
    \centering
    \includegraphics[width=1\linewidth]{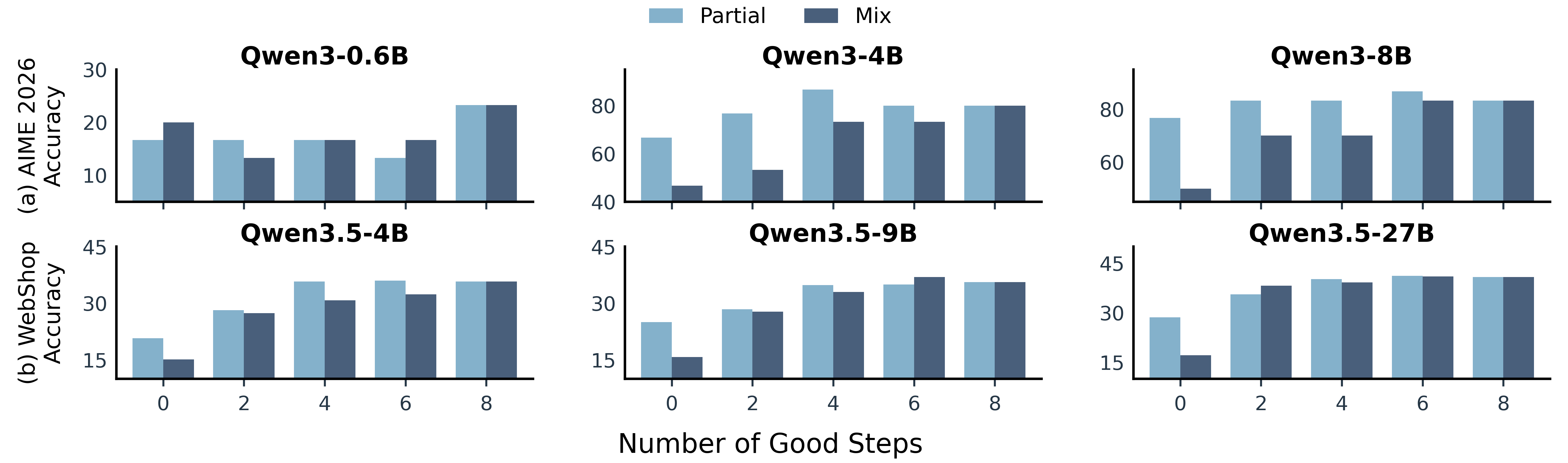}
    \caption{
    Workflow recovery across model scales on \textbf{(a)} AIME 2026 and
    \textbf{(b)} WebShop.
    For each model, Partial retains the first $k$ good steps, whereas Mix combines
    $k$ good steps with $8-k$ bad steps, for $k\in\{0,2,4,6,8\}$.
    Increasing the number of good steps generally narrows the performance gap
    between the two regimes, although recovery can be non-monotonic.
    }
    \label{fig:workflow-recovery-model-scales}
\end{figure}

\textbf{Box$^2$ evaluation design.}
As summarized in Equation~\ref{eq:box2-conditions}, Box$^2$ evaluates each task under five conditions that vary the availability and reliability of workflow guidance. 
No workflow provides the model's independent performance. 
Good and Bad workflows measure its response to reliable versus misleading guidance.
Partial and Mixed workflows share the same useful prefix, so their comparison isolates misleading continuation from the absence of further guidance.

\textbf{Workflow construction and verification.}
A fixed external model serves as a scalable proxy for a human workflow
designer. Using only information available during normal task execution, it
generates a useful workflow
$W^{+}(x)=(g^{+}_1,\ldots,g^{+}_T)$ and, for each change point $k$, a plausible
but misleading continuation
$C_k^{-}(x)=(\tilde g^{-}_{k+1},\ldots,\tilde g^{-}_T)$. These outputs define
\begin{equation}
\begin{alignedat}{2}
W^{G}(x)
&=(g^{+}_1,\ldots,g^{+}_T),
\qquad&
W^{B}(x)
&=C^{-}_0(x)=(\tilde g^{-}_1,\ldots,\tilde g^{-}_T),\\
W^{P}_k(x)
&=(g^{+}_1,\ldots,g^{+}_k),
\qquad&
W^{M}_k(x)
&=(g^{+}_1,\ldots,g^{+}_k,
\tilde g^{-}_{k+1},\ldots,\tilde g^{-}_T).
\end{alignedat}
\end{equation}
The No-workflow condition omits $W(x)$, and $k=0$ yields a workflow that is
misleading. An independent verifier checks useful workflows
for validity and task relevance, misleading workflows for plausibility and
task relevance, and all workflows for answer leakage. Accepted workflows are
then frozen and shared across target models.

\subsection{Evaluation}
\label{sec:box2-revised-evaluation}

\textbf{Benchmarks and models.}
We evaluate Box$^2$ in two complementary settings. The frontier-model evaluation tests Box$^2$ capabilities across diverse tasks, while the open-weight evaluation provides a baseline for our training experiments later.
\textbf{(a) Frontier models.} We evaluate Gemini~3.7~Flash~\citep{google2026gemini37flash}, DeepSeek~V4~Flash~\citep{deepseekai2026deepseekv4}, and GLM~5.2~\citep{glm5team2026glm5} on mathematical reasoning (OpenR1-Math~\citep{benallal2025openr1}), software engineering (DeepSWE~\citep{huang2026deepswe}), information search (BrowseComp~\citep{wei2025browsecomp}), and tool use (AutomationBench~\citep{shepard2026automationbench}).
\textbf{(b) Open-weight models.} We evaluate Qwen3~\citep{yang2025qwen3} and Qwen3.5~\citep{qwen3.5} on AIME~2026~\citep{aime26} and WebShop~\citep{yao2022webshop}, which also serve as the settings for our training experiments. Across both settings, we retain the original tasks, environments, and evaluators, and share each task's frozen workflows across models.

\textbf{Paired workflow effects.}
Let $S_c$ denote the benchmark-level score under condition
$c\in\{0,G,B,P,M\}$. For Partial and Mixed, the score aggregates the
prespecified change points. We define four matched effects,
\begin{equation}
\Delta_{\mathrm{use}}=S_G-S_0,
\qquad
\Delta_{\mathrm{bad}}=S_B-S_0,
\qquad
\Delta_{\mathrm{stop}}=S_P-S_G,
\qquad
\Delta_{\mathrm{switch}}=S_M-S_P.
\end{equation}
$\Delta_{\mathrm{use}}$ measures utilization of useful guidance, while
$\Delta_{\mathrm{bad}}$ measures robustness to guidance that is misleading. $\Delta_{\mathrm{stop}}$ and
$\Delta_{\mathrm{switch}}$ measure recovery when guidance ends or becomes
misleading, respectively. These matched comparisons hold the task, environment,
evaluator, and model fixed, isolating responses to workflow reliability from
absolute task performance.

\subsection{Benchmark Results}
\label{sec:box2-revised-results}

Table~\ref{tab:box2-main-frontier} evaluates how frontier models respond to
workflow guidance when its reliability is fixed or changes during execution.

\textbf{Using reliable guidance and resisting misleading guidance are distinct
capabilities.}
Good workflows improve performance in eight of the twelve frontier
model--task pairs, including gains of $5.9$, $13.7$, and $16.8$ points on
AutomationBench.
Bad workflows, by contrast, reduce performance in all twelve pairs, with
drops ranging from $2.3$ to $43.3$ points.
The open-weight models show the same separation
(Table~\ref{tab:box2-main}): good workflows improve all six model--task pairs,
while bad workflows hurt five of six.
Thus, the ability to benefit from external guidance does not imply the ability
to reject it when it is wrong, revealing a gap between utilization and robustness.

\textbf{Reliance can persist after guidance becomes unreliable.}
Mixed workflows underperform their matched Partial workflows in ten of the
twelve frontier model--task pairs.
Because the two conditions share the same useful prefix and change point, their
difference isolates the effect of continuing with misleading guidance rather
than receiving no further guidance.
The resulting drops show that models often carry forward reliance established
by initially useful workflows instead of revising it when reliability changes
(Figure~\ref{fig:workflow-recovery-model-scales}), revealing inertia in how
reliance is updated.

\textbf{Box$^2$ therefore exposes selective reliance as a capability beyond task
performance.}
Reliable agents must decide not only how to use external guidance, but when
that guidance should continue to influence behavior.
The same limitation appears in open-weight models, motivating us to ask whether
this ability can be learned without sacrificing the benefits of reliable
workflows.

\section{Learning to Think Outside the Box}

\begin{figure}[t]
    \centering
    \includegraphics[width=\linewidth]{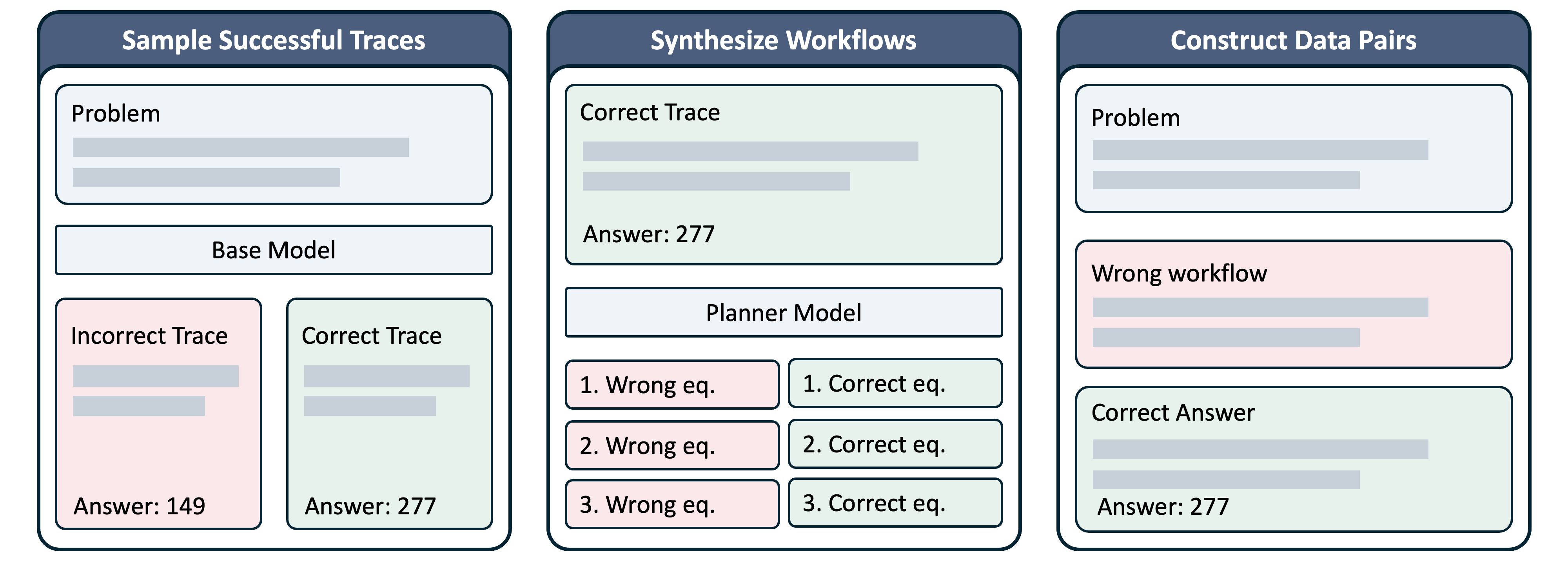}
    \caption{\textbf{Construction of counterfactual supervised fine-tuning data.} For each problem, we sample the base model to obtain a verified successful trace, which the planner uses to synthesize paired correct and incorrect workflows. Each training example combines the problem and an incorrect workflow with the verified correct response, providing supervision for completing the task despite misleading procedural guidance.}
    \label{fig:counterfactual-sft}
\end{figure}

Box$^2$ shows that models can benefit from reliable workflows while remaining
vulnerable to misleading ones. We next ask whether training can improve
robustness without sacrificing utilization. Models receive only bad workflows
during training, while good workflows are reserved for evaluation. Let
$\pi_\theta(\tau\mid x,W)$ denote the model's trajectory distribution for task
$x$ under workflow $W$. The training objectives below constrain this
distribution under $W^B(x)$; behavior under $W^G(x)$ measures generalization
to held-out reliable guidance. Training proceeds through counterfactual
supervised fine-tuning followed by outcome-based reinforcement learning.

\begin{table*}[t]
\centering
\caption{
Performance before and after workflow training on AIME~2026 and WebShop.
$S_P$ and $S_M$ average the three Partial and Mixed conditions, respectively.
Effects report utilization ($S_G-S_0$), robustness ($S_B-S_0$), and recovery
when guidance stops ($S_P-S_G$) or becomes misleading ($S_M-S_P$).
$\mathrm{RL}_{\mathrm{env}}$ uses direct task-outcome rewards on bad-workflow
inputs.
}
\label{tab:training-effects}

\small
\setlength{\tabcolsep}{4.5pt}
\renewcommand{\arraystretch}{1.05}
\begin{tabular}{lcccccdddd}
\toprule

\multirow{3}{*}{\textbf{Model}}
&
\multicolumn{5}{c}{\textbf{Performance}}
&
\multicolumn{4}{c}{\textbf{Box$^2$-Bench Effects}}
\\

\cmidrule(lr){2-6}
\cmidrule(lr){7-10}

&
\textit{Base}
&
\textit{Good}
&
\textit{Bad}
&
\textit{Partial}
&
\textit{Mixed}
&
\cellcolor{effectbg}\textit{Utilization}
&
\cellcolor{effectbg}\textit{Robustness}
&
\multicolumn{2}{c}{\cellcolor{effectbg}\textit{Recovery}}
\\

&
$S_0$
&
$S_G$
&
$S_B$
&
$S_P$
&
$S_M$
&
\cellcolor{effectbg}$\Delta_{\mathrm{use}}$
&
\cellcolor{effectbg}$\Delta_{\mathrm{bad}}$
&
\cellcolor{effectbg}$\Delta_{\mathrm{stop}}$
&
\cellcolor{effectbg}$\Delta_{\mathrm{switch}}$
\\[-1pt]

\midrule
\rowcolor{benchmarkbg}
\multicolumn{10}{l}{
    \rule{0pt}{2ex}
    \small \# \textit{Math --- AIME 2026}
}\\

\textbf{Base}
    & 66.7\%
    & 80.0\%
    & 46.7\%
    & 81.1\%
    & 66.7\%
    & +13.3
    & -20.0
    & +1.1
    & -14.4
\\

\textbf{+ SFT}
    & 73.3\%
    & 70.0\%
    & 66.7\%
    & 74.4\%
    & 70.0\%
    & -3.3
    & -6.7
    & +4.4
    & -4.4
\\

\textbf{+ SFT + $\mathrm{RL}_{\mathrm{env}}$}
    & 73.3\%
    & 80.0\%
    & 56.7\%
    & 73.3\%
    & 74.4\%
    & +6.7
    & -16.7
    & -6.7
    & +1.1
\\

\midrule

\rowcolor{benchmarkbg}
\multicolumn{10}{l}{
    \rule{0pt}{2ex}
    \small \# \textit{Search --- WebShop}
}\\

\textbf{Base}
    & 25.2\%
    & 35.6\%
    & 15.4\%
    & 32.7\%
    & 32.6\%
    & +10.4
    & -9.8
    & -2.9
    & -0.1
\\

\textbf{+ SFT}
    & 30.8\%
    & 33.6\%
    & 30.2\%
    & 32.7\%
    & 31.9\%
    & +2.8
    & -0.6
    & -0.9
    & -0.7
\\

\textbf{+ SFT + $\mathrm{RL}_{\mathrm{env}}$}
    & 32.0\%
    & 34.8\%
    & 30.8\%
    & 33.7\%
    & 33.1\%
    & +2.8
    & -1.2
    & -1.1
    & -0.7
\\

\bottomrule
\end{tabular}
\begin{tablenotes}[flushleft]
\footnotesize
\item[]
\textit{Note.}
Effects measure good-guidance utilization
($\Delta_{\mathrm{use}}=S_G-S_0$),
bad-guidance robustness
($\Delta_{\mathrm{bad}}=S_B-S_0$),
and recovery when guidance stops
($\Delta_{\mathrm{stop}}=S_P-S_G$)
or becomes misleading
($\Delta_{\mathrm{switch}}=S_M-S_P$).
\end{tablenotes}
\end{table*}

\begin{figure}
    \centering
    \includegraphics[width=1\linewidth]{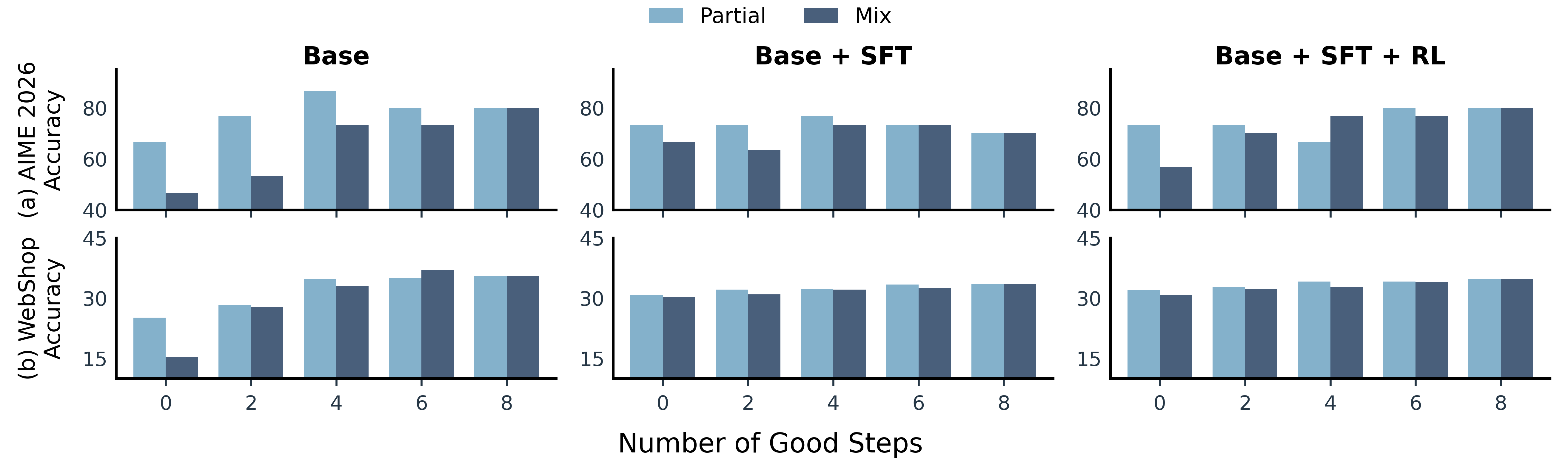}
    \caption{
    Workflow recovery across training stages on \textbf{(a)} AIME 2026 and
    \textbf{(b)} WebShop.
    The columns compare the base model, SFT model, and SFT model followed by RL
    under the same partial- and mixed-workflow compositions.
    SFT generally reduces the separation between the two regimes, while the
    effect of RL depends on the task and workflow composition.
    }
    \label{fig:workflow-recovery-training-stages}
\end{figure}

\subsection{Training with Misleading Workflows}
\label{sec:workflow-training}

\textbf{Counterfactual supervised fine-tuning.}
For each task $x$, we sample trajectories from the base model and retain a
verified successful trajectory $\tau^{+}(x)$. A planner uses this trajectory to
synthesize a useful workflow $W^{G}(x)$ and a plausible but misleading workflow
$W^{B}(x)$. Only the misleading workflow is used for training. 
Figure~\ref{fig:counterfactual-sft} summarizes this three-stage data construction process. 
We minimize the
standard autoregressive negative log-likelihood
\begin{equation}
\mathcal{L}_{\mathrm{SFT}}(\theta)
=
-\mathbb{E}_{x}
\left[
\log \pi_\theta\!\left(\tau^+(x)\mid x,W^B(x)\right)
\right].
\end{equation}
The workflow and target prescribe conflicting strategies, so reducing this
loss increases the probability of successful behavior despite misleading
guidance. For AIME~2026, $\tau^{+}(x)$ is a complete correct solution. For
WebShop, we apply the same construction at each interaction step and use the
successful next action as the target.

This objective trains override behavior without uniquely identifying selective
reliance. Any two policies that agree on bad-workflow inputs obtain the same
SFT loss regardless of how they respond to $W^G(x)$. Consequently, both
selective rejection and broadly ignoring workflows can fit the training data;
preserving held-out utilization must arise through generalization rather than
direct supervision.

\textbf{Outcome-based reinforcement learning.}
Starting from the SFT checkpoint, we continue training on bad-workflow inputs
using task-level outcome rewards. The population objective is
\begin{equation}
J_{\mathrm{RL}}(\theta)
=\mathbb{E}_{x,\,\tau\sim\pi_\theta(\cdot\mid x,W^B(x))}
\!\left[R(x,\tau)\right],
\end{equation}
where $R$ is binary answer correctness for AIME~2026 and the environment
return for WebShop. Thus, $J_{\mathrm{RL}}(\theta)$ is the probability of
success for AIME and the expected task return for WebShop. The reward depends
only on the outcome and never on agreement with the supplied workflow.

For optimization, we sample $G$ trajectories from the current policy snapshot
for each input and compute the group-relative advantage
\begin{equation}
\widehat A_i
=
\frac{R(x,\tau_i)-\operatorname{mean}_{j}R(x,\tau_j)}
{\operatorname{std}_{j}R(x,\tau_j)}.
\end{equation}
We retain only groups with non-degenerate rewards. Let $\rho_{i,t}(\theta)$
denote the likelihood ratio between the current policy and the policy snapshot
that sampled token $t$ of $\tau_i$. We optimize the clipped token-level loss
\begin{equation}
\mathcal{L}_{\mathrm{RL}}(\theta)
=-\mathbb{E}_{i,t}\!\left[
\min\!\left(
\rho_{i,t}(\theta)\widehat A_i,
\operatorname{clip}\!\left(\rho_{i,t}(\theta),
1-\epsilon_{\mathrm{low}},1+\epsilon_{\mathrm{high}}\right)\widehat A_i
\right)
\right]
+\beta D_{\mathrm{KL}}\!\left(\pi_\theta\Vert\pi_{\mathrm{ref}}\right),
\end{equation}
where $\mathbb{E}_{i,t}$ averages over the generated tokens in the retained
groups and $\beta$ controls the optional KL penalty. Unlike SFT, this objective
does not privilege one verified trajectory. It rewards any behavior that
succeeds despite misleading guidance, allowing the model to discover
alternatives to the supervised solution.

\subsection{Training Results}
\label{sec:training-results}

We test whether training on misleading workflows can teach models to override
bad guidance while retaining the ability to use reliable guidance.
We evaluate the base, SFT, and SFT+RL checkpoints under all five Box$^2$
conditions using Qwen3-4B on AIME~2026 and Qwen3.5-9B on
WebShop~\citep{yao2022webshop}.

\textbf{Counterfactual SFT reduces dependence on misleading workflows.}
SFT substantially reduces the damage caused by bad guidance on both tasks.
On AIME, the bad-workflow penalty shrinks from $20.0$ to $6.7$ points; on
WebShop, it shrinks from $9.8$ to $0.6$ points
(Table~\ref{tab:training-effects}).
This robustness is accompanied by weaker reliance on held-out good workflows,
showing that SFT first learns to make workflow guidance defeasible rather than
mandatory.

\textbf{Outcome-based RL recalibrates reliance on guidance.} 
On AIME, RL restores positive utilization of held-out good workflows, raising
$\Delta_{\mathrm{use}}$ from $-3.3$ to $+6.7$, while retaining a robustness improvement over the base model.
On WebShop, the balance established by SFT remains largely stable.
Thus, outcome optimization does not simply reverse SFT; it adjusts the degree
of workflow reliance after robustness has been established.

\textbf{Training also improves adaptation when guidance changes reliability.}
On AIME, the Mixed--Partial gap improves from $-14.4$ for the base model to
$-4.4$ after SFT and $+1.1$ after RL
(Figure~\ref{fig:workflow-recovery-training-stages}).
The trained model can therefore recover when useful guidance becomes
misleading, rather than remaining locked into the earlier workflow.
On WebShop, this gap is already near zero and remains small across training
stages, showing that such recovery is improved or preserved across tasks.

Together, the two stages shape complementary aspects of selective reliance:
SFT teaches models that external guidance can be overridden, while outcome
training restores responsiveness when following guidance is useful,
moving the model toward conditional rather than uniform reliance.

\section{Beyond Workflows}
\label{sec:beyond-workflows}

\begin{figure}[t]
    \centering
    \includegraphics[width=\linewidth]{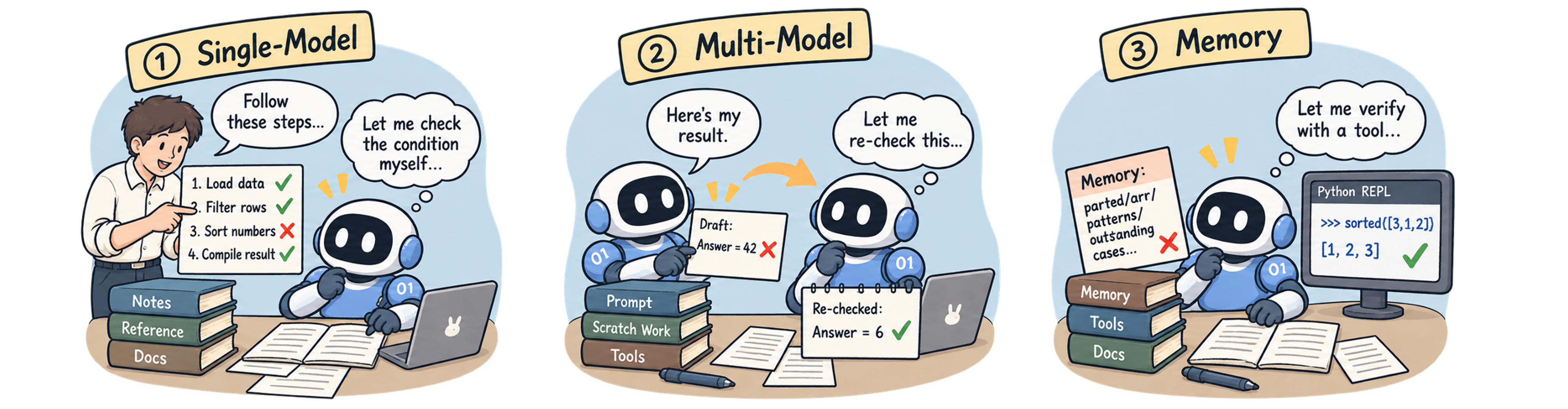}
    \caption{
    \textbf{Selective reliance across sources of context.}
    From left to right, the panels illustrate checking a proposed workflow,
    revising a peer draft, and verifying a stored claim using tool evidence.
    The common challenge is to use helpful context without being bound
    by misleading content.
    Examples and dialogue are schematic rather than recorded trajectories;
    correctness markers are explanatory annotations, not model inputs.
    }
    \label{fig:context_reliance}
\end{figure}

\textbf{Selective reliance across contexts.}
Thinking outside the box also matters in information-rich settings,
where additional context can help or mislead.
Workflow guidance, peer messages, and stored memories provide different
forms of such context (Figure~\ref{fig:context_reliance}).
Across these settings, models must use helpful information while
independently checking questionable content against task constraints
and available evidence.
We therefore evaluate the workflow-trained checkpoints in multi-agent
collaboration and memory-augmented reasoning without further training,
testing whether selective reliance extends beyond procedural guidance.

\textbf{Multi-agent collaboration.}
We test whether workflow training extends to selective use of peer information
in multi-agent reasoning.
We evaluate our checkpoints in the adaptive decentralized environment of
Economy of Minds (EoM)~\citep{qi2026economymindsemergingmultiagent},
where four agents share model weights but maintain separate identities,
memories, and economic states.
For each of 30 AIME~2026 problems, the agents interact for five consecutive
episodes before reset.
We track episode success and peer-induced revisions, distinguishing repairs
(wrong $\rightarrow$ correct) from corruptions (correct $\rightarrow$ wrong).

\begin{figure*}[t]
    \centering

    \begin{subfigure}[t]{0.26\textwidth}
        \centering
        \includegraphics[width=\linewidth]{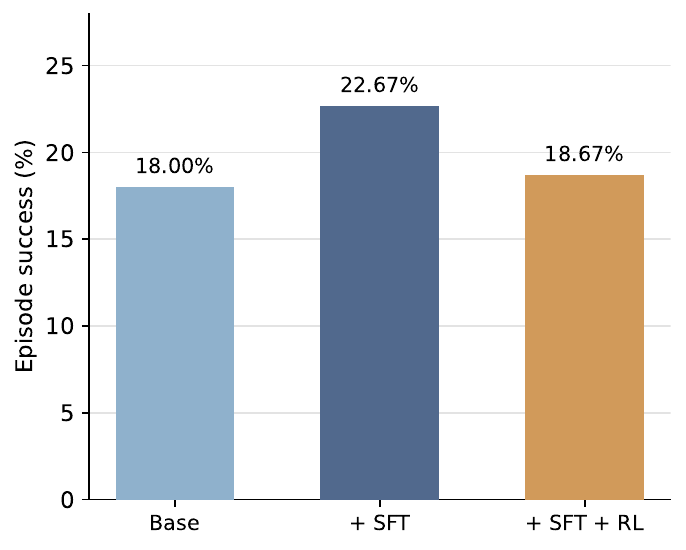}
        \caption{Episode success.}
        \label{fig:eom-a}
    \end{subfigure}\hfill
    \begin{subfigure}[t]{0.44\textwidth}
        \centering
        \includegraphics[width=\linewidth]{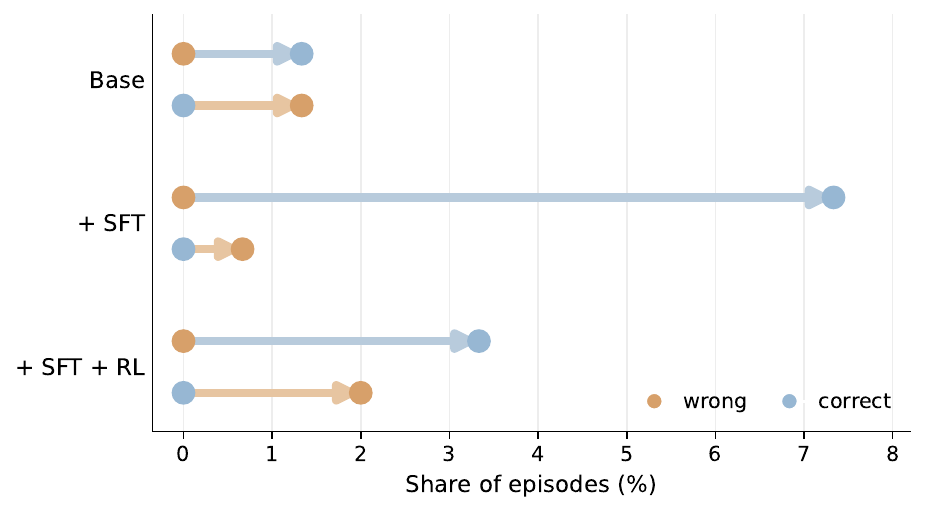}
        \caption{Cross-agent answer revision.}
        \label{fig:eom-b}
    \end{subfigure}\hfill
    \begin{subfigure}[t]{0.26\textwidth}
        \centering
        \includegraphics[width=\linewidth]{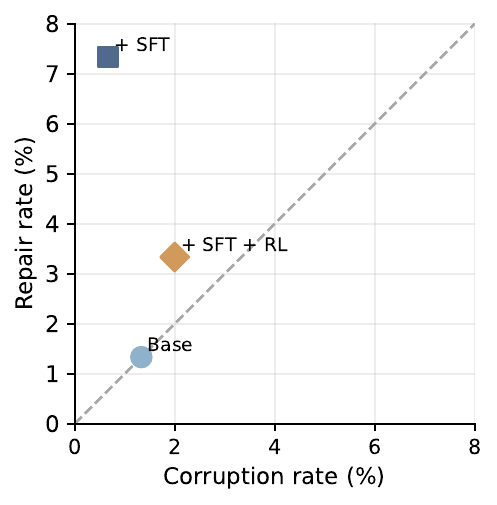}
        \caption{Revision balance.}
        \label{fig:eom-c}
    \end{subfigure}

    \caption{
    Adaptive EoM evaluation on AIME~2026.
    In (b), each model's upper and lower rows show wrong-to-correct and
    correct-to-wrong revisions, respectively; endpoint colors indicate answer
    states, and their horizontal separation gives the corresponding rate.
    The dashed diagonal in (c) marks equal repair and corruption rates.
    All rates are normalized by 150 episodes
    (30 problems $\times$ 5 episodes).
    }
    \label{fig:eom-multiagent}
\end{figure*}

Figure~\ref{fig:eom-multiagent} shows that SFT raises episode success from
18.00\% to 22.67\%, while SFT+RL achieves 18.67\%.
SFT also shifts peer influence toward beneficial revisions, producing 11
repairs and only one corruption, compared with two of each for Base.
SFT+RL produces five repairs and three corruptions.
Overall, our approach extends selective reliance beyond workflows to peer
information, although the gains are not monotonic across training stages.

\begin{table*}[t]
\centering
\caption{
Memory-conditioned performance on LongMemEval-V2-Small.
Score denotes answer accuracy, and Tool Rate denotes the proportion of
trajectories invoking the archive tools; Base denotes no memory brief.
}
\label{tab:memory-transfer}

\small
\setlength{\tabcolsep}{8pt}
\renewcommand{\arraystretch}{1.05}
\begin{tabular}{lcccccc}
\toprule

\multirow{2}{*}{\textbf{Model}}
& \multicolumn{2}{c}{\textbf{Base}}
& \multicolumn{2}{c}{\textbf{Good}}
& \multicolumn{2}{c}{\textbf{Bad}}
\\

\cmidrule(lr){2-3}
\cmidrule(lr){4-5}
\cmidrule(lr){6-7}

& \textit{Score}
& \textit{Tool Rate}
& \textit{Score}
& \textit{Tool Rate}
& \textit{Score}
& \textit{Tool Rate}
\\[-1pt]

\midrule
\rowcolor{benchmarkbg}
\multicolumn{7}{l}{
    \rule{0pt}{2ex}
    \small \# \textit{Memory --- LongMemEval-V2-Small}
}
\\

\textbf{Base}
    & 21.1\%
    & 100.0\%
    & \textbf{95.6\%}
    & 72.2\%
    & 10.0\%
    & 65.6\%
\\

\textbf{+ SFT}
    & \textbf{28.9\%}
    & 100.0\%
    & 92.2\%
    & 72.2\%
    & \textbf{14.4\%}
    & 72.2\%
\\

\textbf{+ SFT + $RL_{\mathrm{env}}$}
    & 25.6\%
    & 100.0\%
    & 94.4\%
    & \textbf{66.7\%}
    & 13.3\%
    & \textbf{74.4\%}
\\

\bottomrule
\end{tabular}

\end{table*}

\textbf{Memory-augmented reasoning.}
We test whether workflow training extends to selective use of stored information.
We evaluate our checkpoints on LongMemEval-V2-Small, following
LongMemEval V2~\citep{wu2026longmemevalv2evaluatinglongtermagent}.
The model receives no memory brief, reliable memory, or corrupted memory while
retaining access to the archive for verification.
We measure both answer accuracy and archive use, which captures whether the
model verifies stored information before relying on it.

Table~\ref{tab:memory-transfer} shows that training preserves the benefit of
reliable memory while improving robustness to corrupted memory.
It also changes verification behavior: the base model consults the archive
less often under corrupted memory than reliable memory, whereas training removes
and eventually reverses this gap.
Our approach extends selective reliance beyond workflows to stored
information, encouraging models to use reliable memory while verifying
potentially misleading records.

\section{Related Work}

\textbf{Frontier Capability and Fallible Procedural Priors.}
LLM agents have progressed from in-context prompting to agent harnesses that structure planning, search, feedback, and tool use~\citep{yao2023react,shinn2023reflexion,yao2023tree,zhou2024lats,liu2025workteam}. A common strategy is to encode task-specific procedural priors through prompts, roles, and workflows, guiding models toward more effective execution \citep{sarukkai2025selfgenerated,zhou2026enhancing,lu2025agent}. This approach is particularly useful when model capability is limited, but frontier agents now exhibit strong performance across a wide range of complex tasks~\citep{huang2026step35,glm5team2026glm5,merrill2026terminalbench,wei2025browsecomp,wijk2025rebench,kwa2025longtasks}. As executor capability increases, we ask whether capable models can benefit from externally designed workflows when they help while remaining robust when they do not provide the best execution path.

\textbf{Harness Generalization across Models and Tasks.}
Harness effectiveness is highly dependent on the model and task: a workflow that helps one executor or setting may provide little benefit, or even become harmful, in another~\citep{yao2026harnessbench,gupta2026noharness,belikova2026proceduralmemory,yu2026compilethenpage,wang2026harnesses}. Existing work therefore focuses largely on adapting the harness itself, through search, repair, or specialization for particular models and deployment settings~\citep{lee2026metaharness,zhang2026selfharness}. Yet in deployment, changes in the executor, task, or environment can leave an existing harness mismatched~\citep{das2025greater,rawles2025androidworld,wang2024agent}. Box$^2$ studies the complementary question: whether the executor itself can adapt, benefiting from useful workflow guidance without being constrained by guidance that no longer fits.

\textbf{Benchmarking Selective Procedural Reliance.}
Recent agent benchmarks measure task completion or procedural compliance.
Outcome-oriented benchmarks evaluate execution across software engineering, web search, terminal use, automation, and research~\citep{huang2026deepswe,wei2025browsecomp,merrill2026terminalbench,shepard2026automationbench,edwards2026rexbench}, while procedure-oriented benchmarks test adherence to instructions, guidelines, and workflows~\citep{qi2025agentif,diao2025guidebench,he2026advancedif,sopbench2026,wang2026sopmaze,jia2026evolif}.
Related work varies guidance quality, instruction priority, or harness configuration~\citep{zhou2026asibench,mccauley2026ihbenchmark,cao2026procedureaware,yao2026harnessbench}, but does not isolate when workflow guidance should be followed or overridden.
Box$^2$-Bench varies workflow validity while holding task and model fixed.
It tests whether models use good guidance and resist bad guidance.

\section{Conclusion}

In this paper, we introduce \emph{thinking outside the box}, the ability to regulate reliance on external guidance, and Box$^2$-Bench, a matched evaluation
that separates this meta-capability from task performance. 
We show that models can benefit from useful workflows yet remain vulnerable to misleading guidance.
We further show that training only on bad workflows can improve robustness, although preserving the benefits of useful guidance remains challenging. 
The resulting behavior also shows preliminary transfer to multi-agent collaboration and memory-augmented reasoning. 
Together, these results suggest that selective reliance is a general ingredient of reliable agent behavior whenever external information is useful but fallible.
Several directions remain open for future study.
(i) We use a fixed external model as a scalable proxy for human workflow designers, and future work should evaluate guidance written by people with more
diverse intentions and error patterns. 
(ii) Box$^2$ varies reliability through controlled and frozen workflows, leaving interactive, ambiguous, partially correct, and model-adaptive guidance to future study.

\clearpage

\bibliography{iclr2027_conference}

@article{wang2026harnesses,
  author = {Wang, Boyuan and Li, Bochao and Wang, Minghan and Tao, Yuxin and Kong, Fang},
  title = {Harnesses for Inference-Time Alignment over Execution Trajectories},
  journal = {arXiv preprint arXiv:2605.21516},
  year = {2026},
  eprint = {2605.21516},
  archivePrefix = {arXiv},
  url = {https://arxiv.org/abs/2605.21516}
}

@inproceedings{yao2023react,
  author = {Yao, Shunyu and Zhao, Jeffrey and Yu, Dian and Du, Nan and Shafran, Izhak and Narasimhan, Karthik R. and Cao, Yuan},
  title = {ReAct: Synergizing Reasoning and Acting in Language Models},
  booktitle = {International Conference on Learning Representations},
  year = {2023},
  url = {https://openreview.net/forum?id=WE_vluYUL-X}
}

@inproceedings{yao2023tree,
  author = {Yao, Shunyu and Yu, Dian and Zhao, Jeffrey and Shafran, Izhak and Griffiths, Tom and Cao, Yuan and Narasimhan, Karthik},
  title = {Tree of Thoughts: Deliberate Problem Solving with Large Language Models},
  booktitle = {Advances in Neural Information Processing Systems},
  volume = {36},
  year = {2023},
  doi = {10.52202/075280-0517},
  url = {https://doi.org/10.52202/075280-0517}
}

@inproceedings{zhou2024lats,
  author = {Zhou, Andy and Yan, Kai and Shlapentokh-Rothman, Michal and Wang, Haohan and Wang, Yu-Xiong},
  title = {Language Agent Tree Search Unifies Reasoning, Acting, and Planning in Language Models},
  booktitle = {Proceedings of the 41st International Conference on Machine Learning},
  series = {Proceedings of Machine Learning Research},
  volume = {235},
  pages = {62138--62160},
  publisher = {PMLR},
  year = {2024},
  url = {https://proceedings.mlr.press/v235/zhou24r.html}
}

@inproceedings{shang2025agentsquare,
  author = {Shang, Yu and Li, Yu and Zhao, Keyu and Ma, Likai and Liu, Jiahe and Xu, Fengli and Li, Yong},
  title = {AgentSquare: Automatic LLM Agent Search in Modular Design Space},
  booktitle = {International Conference on Learning Representations},
  year = {2025},
  url = {https://proceedings.iclr.cc/paper_files/paper/2025/hash/0ae94013da7cd459402fd77874e09ee3-Abstract-Conference.html}
}

@inproceedings{zhang2025aflow,
  author = {Zhang, Jiayi and Xiang, Jinyu and Yu, Zhaoyang and Teng, Fengwei and Chen, Xiong-Hui and Chen, Jiaqi and Zhuge, Mingchen and Cheng, Xin and Hong, Sirui and Wang, Jinlin and Zheng, Bingnan and Liu, Bang and Luo, Yuyu and Wu, Chenglin},
  title = {AFlow: Automating Agentic Workflow Generation},
  booktitle = {International Conference on Learning Representations},
  year = {2025},
  url = {https://openreview.net/forum?id=z5uVAKwmjf}
}

@article{lee2026metaharness,
  author = {Lee, Yoonho and Nair, Roshen and Zhang, Qizheng and Lee, Kangwook and Khattab, Omar and Finn, Chelsea},
  title = {Meta-Harness: End-to-End Optimization of Model Harnesses},
  journal = {arXiv preprint arXiv:2603.28052},
  year = {2026},
  eprint = {2603.28052},
  archivePrefix = {arXiv},
  url = {https://arxiv.org/abs/2603.28052}
}

@article{zhang2026selfharness,
  author = {Zhang, Hangfan and Zhang, Shao and Li, Kangcong and Zhang, Chen and Chen, Yang and Zhang, Yiqun and Bai, Lei and Hu, Shuyue},
  title = {Self-Harness: Harnesses That Improve Themselves},
  journal = {arXiv preprint arXiv:2606.09498},
  year = {2026},
  eprint = {2606.09498},
  archivePrefix = {arXiv},
  url = {https://arxiv.org/abs/2606.09498}
}

@inproceedings{shinn2023reflexion,
  author = {Shinn, Noah and Cassano, Federico and Gopinath, Ashwin and Narasimhan, Karthik and Yao, Shunyu},
  title = {Reflexion: Language Agents with Verbal Reinforcement Learning},
  booktitle = {Advances in Neural Information Processing Systems},
  volume = {36},
  year = {2023},
  doi = {10.52202/075280-0377},
  url = {https://doi.org/10.52202/075280-0377}
}

@inproceedings{wijk2025rebench,
  author = {Wijk, Hjalmar and Lin, Tao Roa and Becker, Joel and Jawhar, Sami and Parikh, Neev and Broadley, Thomas and Chan, Lawrence and Chen, Michael and Clymer, Joshua M. and Dhyani, Jai and Ericheva, Elena and Garcia, Katharyn and Goodrich, Brian and Jurkovic, Nikola and Kinniment, Megan and Lajko, Aron and Nix, Seraphina and Koba Sato, Lucas Jun and Saunders, William and Taran, Maksym and West, Ben and Barnes, Elizabeth},
  title = {{RE}-Bench: Evaluating Frontier {AI} R\&D Capabilities of Language Model Agents against Human Experts},
  booktitle = {Proceedings of the 42nd International Conference on Machine Learning},
  series = {Proceedings of Machine Learning Research},
  volume = {267},
  pages = {66772--66832},
  publisher = {PMLR},
  year = {2025},
  url = {https://proceedings.mlr.press/v267/wijk25a.html}
}

@misc{huang2026step35,
      title={Step 3.5 Flash: Open Frontier-Level Intelligence with 11B Active Parameters}, 
      author={Ailin Huang and Ang Li and Aobo Kong and Bin Wang and Binxing Jiao and Bo Dong and Bojun Wang and Boyu Chen and Brian Li and Buyun Ma and Chang Su and Changxin Miao and Changyi Wan and Chao Lou and Chen Hu and Chen Xu and Chenfeng Yu and Chengting Feng and Chengyuan Yao and Chunrui Han and Dan Ma and Dapeng Shi and Daxin Jiang and Dehua Ma and Deshan Sun and Di Qi and Enle Liu and Fajie Zhang and Fanqi Wan and Guanzhe Huang and Gulin Yan and Guoliang Cao and Guopeng Li and Han Cheng and Hangyu Guo and Hanshan Zhang and Hao Nie and Haonan Jia and Haoran Lv and Hebin Zhou and Hekun Lv and Heng Wang and Heung-Yeung Shum and Hongbo Huang and Hongbo Peng and Hongyu Zhou and Hongyuan Wang and Houyong Chen and Huangxi Zhu and Huimin Wu and Huiyong Guo and Jia Wang and Jian Zhou and Jianjian Sun and Jiaoren Wu and Jiaran Zhang and Jiashu Lv and Jiashuo Liu and Jiayi Fu and Jiayu Liu and Jie Cheng and Jie Luo and Jie Yang and Jie Zhou and Jieyi Hou and Jing Bai and Jingcheng Hu and Jingjing Xie and Jingwei Wu and Jingyang Zhang and Jishi Zhou and Junfeng Liu and Junzhe Lin and Ka Man Lo and Kai Liang and Kaibo Liu and Kaijun Tan and Kaiwen Yan and Kaixiang Li and Kang An and Kangheng Lin and Lei Yang and Liang Lv and Liang Zhao and Liangyu Chen and Lieyu Shi and Liguo Tan and Lin Lin and Lina Chen and Luck Ma and Mengqiang Ren and Michael Li and Ming Li and Mingliang Li and Mingming Zhang and Mingrui Chen and Mitt Huang and Na Wang and Peng Liu and Qi Han and Qian Zhao and Qinglin He and Qinxin Du and Qiuping Wu and Quan Sun and Rongqiu Yang and Ruihang Miao and Ruixin Han and Ruosi Wan and Ruyan Guo and Shan Wang and Shaoliang Pang and Shaowen Yang and Shengjie Fan and Shijie Shang and Shiliang Yang and Shiwei Li and Shuangshuang Tian and Siqi Liu and Siye Wu and Siyu Chen and Song Yuan and Tiancheng Cao and Tianchi Yue and Tianhao Cheng and Tianning Li and Tingdan Luo and Wang You and Wei Ji and Wei Yuan and Wei Zhang and Weibo Wu and Weihao Xie and Wen Sun and Wenjin Deng and Wenzhen Zheng and Wuxun Xie and Xiangfeng Wang and Xiangwen Kong and Xiangyu Liu and Xiangyu Zhang and Xiaobo Yang and Xiaojia Liu and Xiaolan Yuan and Xiaoran Jiao and Xiaoxiao Ren and Xiaoyun Zhang and Xin Li and Xin Liu and Xin Wu and Xing Chen and Xingping Yang and Xinran Wang and Xu Zhao and Xuan He and Xuanti Feng and Xuedan Cai and Xuqiang Zhou and Yanbo Yu and Yang Li and Yang Xu and Yanlin Lai and Yanming Xu and Yaoyu Wang and Yeqing Shen and Yibo Zhu and Yichen Lv and Yicheng Cao and Yifeng Gong and Yijing Yang and Yikun Yang and Yin Zhao and Yingxiu Zhao and Yinmin Zhang and Yitong Zhang and Yixuan Zhang and Yiyang Chen and Yongchi Zhao and Yongshen Long and Yongyao Wang and Yousong Guan and Yu Zhou and Yuang Peng and Yuanhao Ding and Yuantao Fan and Yuanwei Lu and Yuanzhen Yang and Yuchu Luo and Yudi Zhao and Yue Peng and Yueqiang Lin and Yufan Lu and Yuling Zhao and Yunzhou Ju and Yurong Zhang and Yusheng Li and Yuxiang Yang and Yuyang Chen and Yuzhu Cai and Zejia Weng and Zetao Hong and Zexi Li and Zhe Xie and Zheng Ge and Zheng Gong and Zheng Zeng and Zhenyi Lu and Zhewei Huang and Zhichao Chang and Zhiguo Huang and Zhiheng Hu and Zidong Yang and Zili Wang and Ziqi Ren and Zixin Zhang and Zixuan Wang},
      year={2026},
      eprint={2602.10604},
      archivePrefix={arXiv},
      primaryClass={cs.CL},
      url={https://arxiv.org/abs/2602.10604}, 
}

@misc{glm5team2026glm5,
      title={GLM-5: from Vibe Coding to Agentic Engineering}, 
      author={GLM-5-Team and : and Aohan Zeng and Xin Lv and Zhenyu Hou and Zhengxiao Du and Qinkai Zheng and Bin Chen and Da Yin and Chendi Ge and Chenghua Huang and Chengxing Xie and Chenzheng Zhu and Congfeng Yin and Cunxiang Wang and Gengzheng Pan and Hao Zeng and Haoke Zhang and Haoran Wang and Huilong Chen and Jiajie Zhang and Jian Jiao and Jiaqi Guo and Jingsen Wang and Jingzhao Du and Jinzhu Wu and Kedong Wang and Lei Li and Lin Fan and Lucen Zhong and Mingdao Liu and Mingming Zhao and Pengfan Du and Qian Dong and Rui Lu and Shuang-Li and Shulin Cao and Song Liu and Ting Jiang and Xiaodong Chen and Xiaohan Zhang and Xuancheng Huang and Xuezhen Dong and Yabo Xu and Yao Wei and Yifan An and Yilin Niu and Yitong Zhu and Yuanhao Wen and Yukuo Cen and Yushi Bai and Zhongpei Qiao and Zihan Wang and Zikang Wang and Zilin Zhu and Ziqiang Liu and Zixuan Li and Bojie Wang and Bosi Wen and Can Huang and Changpeng Cai and Chao Yu and Chen Li and Chengwei Hu and Chenhui Zhang and Dan Zhang and Daoyan Lin and Dayong Yang and Di Wang and Ding Ai and Erle Zhu and Fangzhou Yi and Feiyu Chen and Guohong Wen and Hailong Sun and Haisha Zhao and Haiyi Hu and Hanchen Zhang and Hanrui Liu and Hanyu Zhang and Hao Peng and Hao Tai and Haobo Zhang and He Liu and Hongwei Wang and Hongxi Yan and Hongyu Ge and Huan Liu and Huanpeng Chu and Jia'ni Zhao and Jiachen Wang and Jiajing Zhao and Jiamin Ren and Jiapeng Wang and Jiaxin Zhang and Jiayi Gui and Jiayue Zhao and Jijie Li and Jing An and Jing Li and Jingwei Yuan and Jinhua Du and Jinxin Liu and Junkai Zhi and Junwen Duan and Kaiyue Zhou and Kangjian Wei and Ke Wang and Keyun Luo and Laiqiang Zhang and Leigang Sha and Liang Xu and Lindong Wu and Lintao Ding and Lu Chen and Minghao Li and Nianyi Lin and Pan Ta and Qiang Zou and Rongjun Song and Ruiqi Yang and Shangqing Tu and Shangtong Yang and Shaoxiang Wu and Shengyan Zhang and Shijie Li and Shuang Li and Shuyi Fan and Wei Qin and Wei Tian and Weining Zhang and Wenbo Yu and Wenjie Liang and Xiang Kuang and Xiangmeng Cheng and Xiangyang Li and Xiaoquan Yan and Xiaowei Hu and Xiaoying Ling and Xing Fan and Xingye Xia and Xinyuan Zhang and Xinze Zhang and Xirui Pan and Xu Zou and Xunkai Zhang and Yadi Liu and Yandong Wu and Yanfu Li and Yidong Wang and Yifan Zhu and Yijun Tan and Yilin Zhou and Yiming Pan and Ying Zhang and Yinpei Su and Yipeng Geng and Yong Yan and Yonglin Tan and Yuean Bi and Yuhan Shen and Yuhao Yang and Yujiang Li and Yunan Liu and Yunqing Wang and Yuntao Li and Yurong Wu and Yutao Zhang and Yuxi Duan and Yuxuan Zhang and Zezhen Liu and Zhengtao Jiang and Zhenhe Yan and Zheyu Zhang and Zhixiang Wei and Zhuo Chen and Zhuoer Feng and Zijun Yao and Ziwei Chai and Ziyuan Wang and Zuzhou Zhang and Bin Xu and Minlie Huang and Hongning Wang and Juanzi Li and Yuxiao Dong and Jie Tang},
      year={2026},
      eprint={2602.15763},
      archivePrefix={arXiv},
      primaryClass={cs.LG},
      url={https://arxiv.org/abs/2602.15763}, 
}

@inproceedings{kwa2025longtasks,
  author = {Kwa, Thomas and West, Ben and Becker, Joel and Deng, Amy and Garcia, Katharyn and Hasin, Max and Jawhar, Sami and Kinniment, Megan and Rush, Nate and Von Arx, Sydney and Bloom, Ryan and Broadley, Thomas and Du, Haoxing and Goodrich, Brian and Jurkovic, Nikola and Miles, Luke and Nix, Seraphina and Lin, Tao and Parikh, Neev and Rein, David and Koba Sato, Lucas Jun and Wijk, Hjalmar and Ziegler, Daniel and Barnes, Elizabeth and Chan, Lawrence},
  title = {Measuring AI Ability to Complete Long Software Tasks},
  booktitle = {Advances in Neural Information Processing Systems},
  volume = {38},
  year = {2025},
  url = {https://proceedings.neurips.cc/paper_files/paper/2025/hash/85069585133c4c168c865e65d72e9775-Abstract-Conference.html}
}

@article{huang2026deepswe,
  author = {Huang, Wenqi and Lee, Charley and Tng, Leonard and Ge, Serena},
  title = {DeepSWE: Measuring Frontier Coding Agents on Original, Long-Horizon Engineering Tasks},
  journal = {arXiv preprint arXiv:2607.07946},
  year = {2026},
  eprint = {2607.07946},
  archivePrefix = {arXiv},
  url = {https://arxiv.org/abs/2607.07946}
}

@article{wei2025browsecomp,
  author = {Wei, Jason and Sun, Zhiqing and Papay, Spencer and McKinney, Scott and Han, Jeffrey and Fulford, Isa and Chung, Hyung Won and Passos, Alex Tachard and Fedus, William and Glaese, Amelia},
  title = {BrowseComp: A Simple Yet Challenging Benchmark for Browsing Agents},
  journal = {arXiv preprint arXiv:2504.12516},
  year = {2025},
  eprint = {2504.12516},
  archivePrefix = {arXiv},
  url = {https://arxiv.org/abs/2504.12516}
}

@inproceedings{
merrill2026terminalbench,
      title={Terminal-Bench: Benchmarking Agents on Hard, Realistic Tasks in Command Line Interfaces}, 
      author={Mike A. Merrill and Alexander G. Shaw and Nicholas Carlini and Boxuan Li and Harsh Raj and Ivan Bercovich and Lin Shi and Jeong Yeon Shin and Thomas Walshe and E. Kelly Buchanan and Junhong Shen and Guanghao Ye and Haowei Lin and Jason Poulos and Maoyu Wang and Marianna Nezhurina and Jenia Jitsev and Di Lu and Orfeas Menis Mastromichalakis and Zhiwei Xu and Zizhao Chen and Yue Liu and Robert Zhang and Leon Liangyu Chen and Anurag Kashyap and Jan-Lucas Uslu and Jeffrey Li and Jianbo Wu and Minghao Yan and Song Bian and Vedang Sharma and Ke Sun and Steven Dillmann and Akshay Anand and Andrew Lanpouthakoun and Bardia Koopah and Changran Hu and Etash Guha and Gabriel H. S. Dreiman and Jiacheng Zhu and Karl Krauth and Li Zhong and Niklas Muennighoff and Robert Amanfu and Shangyin Tan and Shreyas Pimpalgaonkar and Tushar Aggarwal and Xiangning Lin and Xin Lan and Xuandong Zhao and Yiqing Liang and Yuanli Wang and Zilong Wang and Changzhi Zhou and David Heineman and Hange Liu and Harsh Trivedi and John Yang and Junhong Lin and Manish Shetty and Michael Yang and Nabil Omi and Negin Raoof and Shanda Li and Terry Yue Zhuo and Wuwei Lin and Yiwei Dai and Yuxin Wang and Wenhao Chai and Shang Zhou and Dariush Wahdany and Ziyu She and Jiaming Hu and Zhikang Dong and Yuxuan Zhu and Sasha Cui and Ahson Saiyed and Arinbjörn Kolbeinsson and Jesse Hu and Christopher Michael Rytting and Ryan Marten and Yixin Wang and Alex Dimakis and Andy Konwinski and Ludwig Schmidt},
      year={2026},
      eprint={2601.11868},
      archivePrefix={arXiv},
      primaryClass={cs.SE},
      url={https://arxiv.org/abs/2601.11868}, 
}

@article{shepard2026automationbench,
  author = {Shepard, Daniel and Salimans, Robin},
  title = {AutomationBench},
  journal = {arXiv preprint arXiv:2604.18934},
  year = {2026},
  eprint = {2604.18934},
  archivePrefix = {arXiv},
  url = {https://arxiv.org/abs/2604.18934}
}

@article{yao2026harnessbench,
  author = {Yao, Yilun and Tan, Xinyu and Liu, Chao-Hsuan and Li, Yaoming and Wang, Zhengyang and Yu, Wenhan and Tan, Zhewen and Tian, Yuxuan and Zhao, Guangxiang and Sun, Lin and Zhang, Xiangzheng and Yang, Tong},
  title = {Harness-Bench: Measuring Harness Effects across Models in Realistic Agent Workflows},
  journal = {arXiv preprint arXiv:2605.27922},
  year = {2026},
  eprint = {2605.27922},
  archivePrefix = {arXiv},
  url = {https://arxiv.org/abs/2605.27922}
}

@article{gupta2026noharness,
  author = {Gupta, Akshat and Lei, Jermaine and Lu, Alexander and Anumanchipalli, Gopala and Choshen, Leshem},
  title = {Automated Discovery Has No Universally Superior Harness},
  journal = {arXiv preprint arXiv:2607.18235},
  year = {2026},
  eprint = {2607.18235},
  archivePrefix = {arXiv},
  url = {https://arxiv.org/abs/2607.18235}
}

@article{belikova2026proceduralmemory,
  author = {Belikova, Julia and Parchiev, Rauf and Egorov, Evgeny and Davydenko, Grigorii and Gusev, Gleb and Savchenko, Andrey and Makarenko, Maksim},
  title = {Managing Procedural Memory in LLM Agents: Control, Adaptation, and Evaluation},
  journal = {arXiv preprint arXiv:2606.23127},
  year = {2026},
  eprint = {2606.23127},
  archivePrefix = {arXiv},
  url = {https://arxiv.org/abs/2606.23127}
}

@article{yu2026compilethenpage,
  author = {Yu, Chenglin and Yin, Li and Yu, Ying and Yang, Hongxia and Li, Ming},
  title = {Compile, Then Page: Executable SOP Programs and a Capability-Gated Runtime for Procedural LLM Agents},
  journal = {arXiv preprint arXiv:2607.11346},
  year = {2026},
  eprint = {2607.11346},
  archivePrefix = {arXiv},
  url = {https://arxiv.org/abs/2607.11346}
}

@inproceedings{edwards2026rexbench,
  author = {Edwards, Nicholas and Lee, Yukyung and Mao, Yujun Audrey and Qin, Yulu and Schuster, Sebastian and Kim, Najoung},
  title = {RExBench: Can coding agents autonomously implement AI research extensions?},
  booktitle = {Proceedings of the 64th Annual Meeting of the Association for Computational Linguistics (Volume 1: Long Papers)},
  pages = {16380--16417},
  publisher = {Association for Computational Linguistics},
  year = {2026},
  doi = {10.18653/v1/2026.acl-long.745},
  url = {https://aclanthology.org/2026.acl-long.745/}
}

@inproceedings{qi2025agentif,
  author = {Qi, Yunjia and Peng, Hao and Wang, Xiaozhi and Xin, Amy and Liu, Youfeng and Xu, Bin and Hou, Lei and Li, Juanzi},
  title = {AGENTIF: Benchmarking Large Language Models Instruction Following Ability in Agentic Scenarios},
  booktitle = {Advances in Neural Information Processing Systems},
  volume = {38},
  year = {2025},
  url = {https://proceedings.neurips.cc/paper_files/paper/2025/hash/51bb3a8a33610a25aae074bfc51b1b1f-Abstract-Datasets_and_Benchmarks_Track.html}
}

@inproceedings{diao2025guidebench,
  author = {Diao, Lingxiao and Xu, Xinyue and Sun, Wanxuan and Yang, Cheng and Zhang, Zhuosheng},
  title = {GuideBench: Benchmarking Domain-Oriented Guideline Following for LLM Agents},
  booktitle = {Proceedings of the 63rd Annual Meeting of the Association for Computational Linguistics (Volume 1: Long Papers)},
  pages = {11361--11399},
  publisher = {Association for Computational Linguistics},
  year = {2025},
  doi = {10.18653/v1/2025.acl-long.557},
  url = {https://aclanthology.org/2025.acl-long.557/}
}

@inproceedings{he2026advancedif,
  author = {He, Yun and Li, Wenzhe and Zhang, Hejia and Li, Songlin and Mandyam, Karishma and Khosla, Sopan and Xiong, Yuanhao and Wang, Nanshu and Peng, Xiaoliang and Li, Beibin and Bi, Shengjie and Patil, Shishir G. and Qi, Qi and Feng, Shengyu and Katz-Samuels, Julian and Pang, Richard Yuanzhe and Gonugondla, Sujan Kumar and Lang, Hunter and Yu, Yue and Qian, Yundi and Fazel-Zarandi, Maryam and Yu, Licheng and Benhalloum, Amine and Awadalla, Hany Hassan and Faruqui, Manaal},
  title = {AdvancedIF: Rubric-Based Benchmarking and Reinforcement Learning for Advancing LLM Instruction Following},
  booktitle = {Proceedings of the 64th Annual Meeting of the Association for Computational Linguistics (Volume 1: Long Papers)},
  pages = {18003--18022},
  year = {2026},
  doi = {10.18653/v1/2026.acl-long.820},
  url = {https://aclanthology.org/2026.acl-long.820/}
}

@inproceedings{sopbench2026,
  title={Sop-bench: Complex industrial sops for evaluating llm agents},
  author={Nandi, Subhrangshu and Datta, Arghya and Nama, Rohith and Patel, Udita and Vichare, Nikhil and Bhattacharya, Indranil and Asija, Shivam and Gupta, Arushi and Carenini, Giuseppe and Xu, Jing and others},
  booktitle={Proceedings of the 32nd ACM SIGKDD Conference on Knowledge Discovery and Data Mining V. 2},
  pages={9604--9615},
  year={2026}
}

@inproceedings{wang2026sopmaze,
  author = {Wang, Jiaming and Tang, Zhe and Jin, Zehao and Chen, Hefei and Jin, Yilin and Ding, Peng and Li, Xiaoyu and Cao, Xuezhi},
  title = {SOP-Maze: Evaluating Large Language Models on Complicated Business Standard Operating Procedures},
  booktitle = {Findings of the Association for Computational Linguistics: ACL 2026},
  pages = {14568--14588},
  publisher = {Association for Computational Linguistics},
  year = {2026},
  doi = {10.18653/v1/2026.findings-acl.715},
  url = {https://aclanthology.org/2026.findings-acl.715/}
}

@inproceedings{jia2026evolif,
  author = {Jia, Qi and Shen, Ye and Song, Xiujie and Zhang, Kaiwei and Wang, Shibo and Pei, Dun and Zhu, Xiangyang and Zhai, Guangtao},
  title = {One Battle After Another: Probing LLMs' Limits on Multi-Turn Instruction Following with a Benchmark Evolving Framework},
  booktitle = {Proceedings of the 64th Annual Meeting of the Association for Computational Linguistics (Volume 1: Long Papers)},
  pages = {9574--9590},
  publisher = {Association for Computational Linguistics},
  year = {2026},
  doi = {10.18653/v1/2026.acl-long.433},
  url = {https://aclanthology.org/2026.acl-long.433/}
}

@article{zhou2026asibench,
  author = {Zhou, Junwei and Sun, Zhen and Li, Binyu and Zhou, Jiangyu and Pan, Yuexi and Wang, Hengyu and Ren, Honghe and Jia, Xiaohan and Zhou, Xueyang and Cao, Xiaoyu and Chen, Yongchao and Feng, Yuanning and Wu, Junhao and Zhang, Cheng and Chen, Sijia and Xue, Haoyu and You, Chengsong and Wang, Huan and Wu, Koutian and Gao, Peigan and Wu, Jiakun and Li, Wenzhe and Shang, Ergan and Zheng, Qingyuan and Zhou, Jingjing and Jia, Ruixuan and Xu, Yan and Zhang, Hongrui and Ma, Xiao-Han and Cheng, Zhengxiang and Hao, Yuexing and Mai, Liting and Ji, Xianglin and Zhang, Wenjun and Chen, Zhuofan and Huang, Yixiao and Wang, Chi and Hua, Wenyue and Hao, Yilun and Zhai, Yuantao and Zhao, Ziyan and Xie, Jingyan},
  title = {ASI-Bench: At the Dawn of Artificial Superintelligence},
  journal = {arXiv preprint arXiv:2608.17271},
  year = {2026},
  eprint = {2608.17271},
  archivePrefix = {arXiv},
  url = {https://arxiv.org/abs/2608.17271}
}

@article{mccauley2026ihbenchmark,
  author = {McCauley, Conor and Kan, Zeliang and Martin, Jason},
  title = {IH-Benchmark: A Conflict-Centered Benchmark for Instruction-Hierarchy Robustness in LLM Applications},
  journal = {arXiv preprint arXiv:2607.25987},
  year = {2026},
  eprint = {2607.25987},
  archivePrefix = {arXiv},
  url = {https://arxiv.org/abs/2607.25987}
}

@article{cao2026procedureaware,
  author = {Cao, Hongliu and Driouich, Ilias and Thomas, Eoin},
  title = {Beyond Task Completion: Revealing Corrupt Success in LLM Agents through Procedure-Aware Evaluation},
  journal = {arXiv preprint arXiv:2603.03116},
  year = {2026},
  eprint = {2603.03116},
  archivePrefix = {arXiv},
  url = {https://arxiv.org/abs/2603.03116}
}

@inproceedings{liu2025workteam,
  author = {Liu, Hanchao and Li, Rongjun and Xiong, Weimin and Zhou, Ziyu and Peng, Wei},
  title = {WorkTeam: Constructing Workflows from Natural Language with Multi-Agents},
  booktitle = {Proceedings of the 2025 Conference of the Nations of the Americas Chapter of the Association for Computational Linguistics: Human Language Technologies (Volume 3: Industry Track)},
  pages = {20--35},
  publisher = {Association for Computational Linguistics},
  year = {2025},
  doi = {10.18653/v1/2025.naacl-industry.3},
  url = {https://aclanthology.org/2025.naacl-industry.3/}
}

@inproceedings{sarukkai2025selfgenerated,
  author = {Sarukkai, Vishnu and Xie, Zhiqiang and Fatahalian, Kayvon},
  title = {Self-Generated In-Context Examples Improve LLM Agents for Sequential Decision-Making Tasks},
  booktitle = {Advances in Neural Information Processing Systems},
  volume = {38},
  year = {2025},
  doi = {10.52202/085713-2158},
  url = {https://proceedings.neurips.cc/paper_files/paper/2025/hash/5d1f02132ef51602adf07000ca5b6138-Abstract-Conference.html}
}

@misc{wu2026longmemevalv2evaluatinglongtermagent,
  author = {Di Wu and Zixiang Ji and Asmi Kawatkar and Bryan Kwan and Jia-Chen Gu and Nanyun Peng and Kai-Wei Chang},
  title = {LongMemEval-V2: Evaluating Long-Term Agent Memory Toward Experienced Colleagues},
  journal = {arXiv preprint arXiv:2605.12493},
  year = {2026},
  eprint = {2605.12493},
  archivePrefix = {arXiv},
  primaryClass = {cs.CL},
  url = {https://arxiv.org/abs/2605.12493}
}

@inproceedings{zhou2026enhancing,
 author = {Zhou, Jiacong and Miao, Jiaxu and wang, xianyun and Yu, Jun},
 booktitle = {Advances in Neural Information Processing Systems},
 doi = {10.52202/085713-4246},
 editor = {D. Belgrave and C. Zhang and H. Lin and R. Pascanu and P. Koniusz and M. Ghassemi and N. Chen},
 pages = {127466--127495},
 publisher = {Curran Associates, Inc.},
 title = {Enhancing LLM Planning for Robotics Manipulation through Hierarchical Procedural Knowledge Graphs},
 url = {https://proceedings.neurips.cc/paper_files/paper/2025/file/b94310e1c7ecb79f1a24adc757f1b89b-Paper-Conference.pdf},
 volume = {38, Main Conference},
 year = {2025}
}

@inproceedings{lu2025agent,
  author = {Lu, Kai and Xu, Shixiong and Li, Jinqiu and Ding, Kun and Meng, Gaofeng},
  title = {Agent Reviewers: Domain-specific Multimodal Agents with Shared Memory for Paper Review},
  booktitle = {Proceedings of the 42nd International Conference on Machine Learning},
  series = {Proceedings of Machine Learning Research},
  volume = {267},
  pages = {40803--40830},
  publisher = {PMLR},
  year = {2025},
  url = {https://proceedings.mlr.press/v267/lu25p.html}
}

@inproceedings{das2025greater,
  author = {Das, Sarkar Snigdha Sarathi and Kamoi, Ryo and Pang, Bo and Zhang, Yusen and Xiong, Caiming and Zhang, Rui},
  title = {{GReaTer}: Gradients Over Reasoning Makes Smaller Language Models Strong Prompt Optimizers},
  booktitle = {International Conference on Learning Representations},
  year = {2025},
  url = {https://proceedings.iclr.cc/paper_files/paper/2025/hash/18a42aad2fa8aa871e2ee20d425c208d-Abstract-Conference.html}
}

@inproceedings{rawles2025androidworld,
  author = {Rawles, Chris and Clinckemaillie, Sarah and Chang, Yifan and Waltz, Jonathan and Lau, Gabrielle and Fair, Marybeth and Li, Alice and Bishop, William and Li, Wei and Campbell-Ajala, Folawiyo and Toyama, Daniel and Berry, Robert and Tyamagundlu, Divya and Lillicrap, Timothy and Riva, Oriana},
  title = {{AndroidWorld}: A Dynamic Benchmarking Environment for Autonomous Agents},
  booktitle = {International Conference on Learning Representations},
  year = {2025},
  url = {https://proceedings.iclr.cc/paper_files/paper/2025/hash/01a83bc2f2732a58e6aa731e659e7101-Abstract-Conference.html}
}

@article{wang2024agent,
  author = {Wang, Zora Zhiruo and Mao, Jiayuan and Fried, Daniel and Neubig, Graham},
  title = {Agent workflow memory},
  journal = {arXiv preprint arXiv:2409.07429},
  year = {2024},
  eprint = {2409.07429},
  archivePrefix = {arXiv},
  url = {https://arxiv.org/abs/2409.07429}
}

@misc{li2026flow,
      title={In-the-Flow Agentic System Optimization for Effective Planning and Tool Use}, 
      author={Zhuofeng Li and Haoxiang Zhang and Seungju Han and Sheng Liu and Jianwen Xie and Yu Zhang and Yejin Choi and James Zou and Pan Lu},
      year={2026},
      eprint={2510.05592},
      archivePrefix={arXiv},
      primaryClass={cs.AI},
      url={https://arxiv.org/abs/2510.05592}, 
}

@article{ruan2026aorchestra,
  author = {Ruan, Jianhao and Xu, Zhihao and Peng, Yiran and Ren, Fashen and Yu, Zhaoyang and Liang, Xinbing and Xiang, Jinyu and Liu, Bang and Wu, Chenglin and Luo, Yuyu and Zhang, Jiayi},
  title = {{AOrchestra}: Automating Sub-Agent Creation for Agentic Orchestration},
  journal = {arXiv preprint arXiv:2602.03786},
  year = {2026},
  eprint = {2602.03786},
  archivePrefix = {arXiv},
  url = {https://arxiv.org/abs/2602.03786}
}

@article{feng2026towards,
  author = {Feng, Tony and Trinh, Trieu H. and Bingham, Garrett and Hwang, Dawsen and Chervonyi, Yuri and Jung, Junehyuk and Lee, Joonkyung and Pagano, Carlo and Kim, Sang-hyun and Pasqualotto, Federico and Gukov, Sergei and Lee, Jonathan N. and Kim, Junsu and Hou, Kaiying and Ghiasi, Golnaz and Tay, Yi and Li, YaGuang and Kuang, Chenkai and Liu, Yuan and Lin, Hanzhao and Liu, Evan Zheran and Nayakanti, Nigamaa and Yang, Xiaomeng and Cheng, Heng-Tze and Hassabis, Demis and Kavukcuoglu, Koray and Le, Quoc V. and Luong, Thang},
  title = {Towards Autonomous Mathematics Research},
  journal = {arXiv preprint arXiv:2602.10177},
  year = {2026},
  eprint = {2602.10177},
  archivePrefix = {arXiv},
  url = {https://arxiv.org/abs/2602.10177}
}

@article{kung2026leap,
  author = {Kung, Po-Nien and Song, Linfeng and Hwang, Dawsen and Yoon, Jinsung and Li, Chun-Liang and Severini, Simone and Ol{\v{s}}{\'a}k, Mirek and Lockhart, Edward and Le, Quoc V. and Gokturk, Burak and Luong, Thang and Pfister, Tomas and Peng, Nanyun},
  title = {{LEAP}: Supercharging {LLM}s for Formal Mathematics with Agentic Frameworks},
  journal = {arXiv preprint arXiv:2606.03303},
  year = {2026},
  eprint = {2606.03303},
  archivePrefix = {arXiv},
  url = {https://arxiv.org/abs/2606.03303}
}

@article{yamada2026towards,
  author = {Lu, Chris and Lu, Cong and Lange, Robert Tjarko and Yamada, Yutaro and Hu, Shengran and Foerster, Jakob and Ha, David and Clune, Jeff},
  title = {Towards end-to-end automation of {AI} research},
  journal = {Nature},
  volume = {651},
  pages = {914--919},
  year = {2026},
  doi = {10.1038/s41586-026-10265-5},
  url = {https://www.nature.com/articles/s41586-026-10265-5}
}

@article{ma2026lower,
  author = {Ma, Jianhao and Chen, Yuxin},
  title = {A lower bound for stepsize-based acceleration of gradient descent},
  journal = {arXiv preprint arXiv:2608.10418},
  year = {2026},
  eprint = {2608.10418},
  archivePrefix = {arXiv},
  url = {https://arxiv.org/abs/2608.10418}
}

@inproceedings{li2026agencybench,
  author = {Li, Keyu and Shi, Junhao and Xiao, Yang and Jiang, Mohan and Sun, Jie and Wu, Yunze and Fu, Dayuan and Xia, Shijie and Cai, Xiaojie and Xu, Tianze and Si, Weiye and Li, Wenjie and Wang, Dequan and Liu, Pengfei},
  title = {{AgencyBench}: Benchmarking the Frontiers of Autonomous Agents in 1M-Token Real-World Contexts},
  booktitle = {Proceedings of the 64th Annual Meeting of the Association for Computational Linguistics (Volume 1: Long Papers)},
  pages = {7422--7440},
  publisher = {Association for Computational Linguistics},
  year = {2026},
  doi = {10.18653/v1/2026.acl-long.337},
  url = {https://aclanthology.org/2026.acl-long.337/}
}

@article{deng2025swe,
  author = {Deng, Xiang and Da, Jeff and Pan, Edwin and He, Yannis Yiming and Ide, Charles and Garg, Kanak and Lauffer, Niklas and Park, Andrew and Pasari, Nitin and Rane, Chetan and Sampath, Karmini and Krishnan, Maya and Kundurthy, Srivatsa and Hendryx, Sean and Wang, Zifan and Zhang, Chen Bo Calvin and Jacobson, Noah and Liu, Bing and Kenstler, Brad},
  title = {{SWE-Bench Pro}: Can {AI} Agents Solve Long-Horizon Software Engineering Tasks?},
  journal = {arXiv preprint arXiv:2509.16941},
  year = {2025},
  eprint = {2509.16941},
  archivePrefix = {arXiv},
  url = {https://arxiv.org/abs/2509.16941}
}

@inproceedings{guan2026supchain,
  author = {Guan, Shengyue and Liu, Yihao and Cao, Lang},
  title = {SupChain-Bench: Benchmarking Large Language Models for Real-World Supply Chain Management},
  booktitle = {Findings of the Association for Computational Linguistics: ACL 2026},
  pages = {7526--7550},
  publisher = {Association for Computational Linguistics},
  year = {2026},
  doi = {10.18653/v1/2026.findings-acl.371},
  url = {https://aclanthology.org/2026.findings-acl.371/}
}

@misc{qi2026economymindsemergingmultiagent,
  author = {Zhenting Qi and Huangyuan Su and Ao Qu and Chenyu Wang and Yu Yao and Han Zheng and Kushal Chattopadhyay and Guowei Xu and Zihan Wang and Weirui Ye and Vijay Janapa Reddi and Ju Li and Paul Pu Liang and Himabindu Lakkaraju and Sham Kakade and Yilun Du},
  title = {Economy of Minds: Emerging Multi-Agent Intelligence with Economic Interactions},
  journal = {arXiv preprint arXiv:2606.02859},
  year = {2026},
  eprint = {2606.02859},
  archivePrefix = {arXiv},
  primaryClass = {cs.CL},
  url = {https://arxiv.org/abs/2606.02859}
}

@inproceedings{yao2022webshop,
 author = {Yao, Shunyu and Chen, Howard and Yang, John and Narasimhan, Karthik},
 booktitle = {Advances in Neural Information Processing Systems},
 doi = {10.52202/068431-1508},
 editor = {S. Koyejo and S. Mohamed and A. Agarwal and D. Belgrave and K. Cho and A. Oh},
 pages = {20744--20757},
 publisher = {Curran Associates, Inc.},
 title = {WebShop: Towards Scalable Real-World Web Interaction with Grounded Language Agents},
 url = {https://proceedings.neurips.cc/paper_files/paper/2022/file/82ad13ec01f9fe44c01cb91814fd7b8c-Paper-Conference.pdf},
 volume = {35},
 year = {2022}
}

@misc{google2026gemini37flash,
  title={Gemini 3.7 Flash Model Card},
  author={{Google DeepMind}},
  year={2026},
  howpublished={\url{https://deepmind.google/models/model-cards/gemini-3-7-flash}},
  note={Accessed 2026-08-13}
}

@misc{deepseekai2026deepseekv4,
  title         = {DeepSeek-V4: Towards Highly Efficient Million-Token Context Intelligence},
  author        = {{DeepSeek-AI}},
  year          = {2026},
  eprint        = {2606.19348},
  archivePrefix = {arXiv},
  primaryClass  = {cs.CL},
  url           = {https://arxiv.org/abs/2606.19348}
}

@misc{benallal2025openr1,
  title        = {Open R1: Update \#2},
  author       = {Ben Allal, Loubna and Tunstall, Lewis and Lozhkov, Anton and Bakouch, Elie and Penedo, Guilherme and Kydlicek, Hynek and Mart{\'i}n Bl{\'a}zquez, Gabriel},
  year         = {2025},
  month        = feb,
  howpublished = {Hugging Face Blog},
  url          = {https://huggingface.co/blog/open-r1/update-2}
}

@misc{yang2025qwen3,
  title         = {Qwen3 Technical Report},
  author        = {Yang, An and Li, Anfeng and Yang, Baosong and Zhang, Beichen and Hui, Binyuan and Zheng, Bo and Yu, Bowen and Gao, Chang and Huang, Chengen and Lv, Chenxu and Zheng, Chujie and Liu, Dayiheng and Zhou, Fan and Huang, Fei and Hu, Feng and Ge, Hao and Wei, Haoran and Lin, Huan and Tang, Jialong and Yang, Jian and Tu, Jianhong and Zhang, Jianwei and Yang, Jianxin and Yang, Jiaxi and Zhou, Jing and Zhou, Jingren and Lin, Junyang and Dang, Kai and Bao, Keqin and Yang, Kexin and Yu, Le and Deng, Lianghao and Li, Mei and Xue, Mingfeng and Li, Mingze and Zhang, Pei and Wang, Peng and Zhu, Qin and Men, Rui and Gao, Ruize and Liu, Shixuan and Luo, Shuang and Li, Tianhao and Tang, Tianyi and Yin, Wenbiao and Ren, Xingzhang and Wang, Xinyu and Zhang, Xinyu and Ren, Xuancheng and Fan, Yang and Su, Yang and Zhang, Yichang and Zhang, Yinger and Wan, Yu and Liu, Yuqiong and Wang, Zekun and Cui, Zeyu and Zhang, Zhenru and Zhou, Zhipeng and Qiu, Zihan},
  year          = {2025},
  eprint        = {2505.09388},
  archivePrefix = {arXiv},
  primaryClass  = {cs.CL},
  url           = {https://arxiv.org/abs/2505.09388}
}

@misc{qwen3.5,
  title  = {{Qwen3.5}: Towards Native Multimodal Agents},
  author = {{Qwen Team}},
  month  = feb,
  year   = {2026},
  url    = {https://qwen.ai/blog?id=qwen3.5}
}

@misc{aime26,
  title  = {American Invitational Mathematics Examination (AIME) 2026},
  author = {Zhang, Yifan and Math-AI, Team},
  year   = {2026}
}

@inproceedings{yang2026int,
 author = {Yang, Matthew and Bai, Hao and Wu, Ian and Yang, Gene and Setlur, Amrith and Kumar, Aviral},
 booktitle = {International Conference on Learning Representations},
 editor = {C. Vondrick and B. Hariharan and C. Raffel and L. Pinto and D. Yang and A. Faust},
 pages = {85054--85091},
 title = {InT: Self-Proposed Interventions Enable Credit Assignment in LLM Reasoning},
 url = {https://proceedings.iclr.cc/paper_files/paper/2026/file/89062e4d480c0c3a88d36c20c5694459-Paper-Conference.pdf},
 volume = {2026},
 year = {2026}
}
\bibliographystyle{iclr2027_conference}

\clearpage
\appendix
\section{Experimental Details}
\label{app:experimental-settings}

\paragraph{Released resources.}
Box$^2$-Bench and the trained model checkpoints are publicly available
through our
\href{https://huggingface.co/collections/ElvisWang111/thinking-outside-the-box}
{Hugging Face collection}.

\subsection{Training Data Construction}
\label{app:training-data}

We build separate training sets for mathematical reasoning and WebShop. In
each domain, reinforcement learning draws its tasks from the supervised
fine-tuning pool. The example and task counts differ because a mathematical
problem may have several correct solutions, whereas each action in a WebShop
trajectory becomes one supervised example.

\subsubsection{Mathematical Reasoning}

\paragraph{Source problems.}
We use 2,060 problems from the InT-SFT \citep{yang2026int} training split. For
each problem, we sample Qwen3-8B sixteen times and keep solutions whose
extracted final answers match the reference answer.

\paragraph{Workflow construction.}
For each problem, we ask Qwen3.5-Plus-02-15 to generate a pair of five-step
workflows from the problem statement and verified Qwen3-8B solution traces.
The \emph{good workflow} summarizes a supported solution strategy. The
\emph{bad workflow} develops a coherent, incorrect reasoning path from an
explicit failure mode. We keep only pairs in which both workflows have exactly
five steps and neither reveals the final answer.

\paragraph{Supervised fine-tuning.}
Each SFT example pairs a problem and its bad workflow with a verified correct
solution,
\begin{equation}
x_{\mathrm{SFT}}=(\text{problem},\text{bad workflow}),
\qquad
y_{\mathrm{SFT}}=\text{verified correct solution}.
\end{equation}
For each retained problem, we pair its bad workflow with every distinct
verified correct solution available for that problem. After filtering, this
procedure produces 2,555 prompt--completion pairs.

\paragraph{Reinforcement learning.}
Math RL uses the SFT problem pool and supplies only bad workflows. We draw
sixteen rollouts per prompt from the SFT checkpoint and keep prompts with one
to fifteen correct rollouts. These bounds remove groups with constant binary
reward and leave 111 bad-workflow prompts. A completion receives reward one
when its extracted final answer matches the reference answer and zero
otherwise.

\subsubsection{WebShop}

\paragraph{Source tasks.}
We use only the official WebShop training split, whose global task indices
range from 1500 to 12086 \citep{yao2022webshop}.

\paragraph{Workflow construction.}
For each selected training task, we ask Qwen3.5-Plus-02-15 to generate a pair
of eight-step workflows from the shopping instruction and a successful action
trajectory. The \emph{good workflow} summarizes the successful purchase
strategy. The \emph{bad workflow} follows a coherent, incorrect strategy that
violates explicit task constraints. After structural filtering, this procedure
produces 2,023 workflow pairs.

\paragraph{Supervised fine-tuning.}
WebShop SFT uses only the bad workflow. At each turn, the model receives the
task instruction, the task-level bad workflow, the preceding interaction
history, the current observation, and the legal actions. The target is the
successful next action,
\begin{equation}
\begin{aligned}
x_{\mathrm{SFT}}
&=(\text{task},\text{bad workflow},\text{history},
\text{observation},\text{legal actions}),\\
y_{\mathrm{SFT}}
&=\text{successful next action}.
\end{aligned}
\end{equation}
Expanding the 2,023 trajectories into turn-level supervision produces 8,203
training examples.

\paragraph{Reinforcement learning.}
WebShop RL draws tasks from the SFT pool and pairs each task with its bad
workflow. To identify tasks that provide within-group reward variation for
Group Relative Policy Optimization (GRPO), we generate eight interactive rollouts per task using the SFT checkpoint
and retain tasks whose terminal rewards are not all equal. This filtering
leaves 153 tasks. A deterministic split assigns 139 tasks to RL training and
14 tasks to an in-domain monitoring set. Each training rollout uses an
independent WebShop session and receives the terminal environment reward in
$[0,1]$.

\subsection{Evaluation Protocol and Data Separation}
\label{app:evaluation-protocol}

All optimization that updates model parameters uses only training tasks. We
evaluate mathematical reasoning on all 30 AIME~2026 problems and WebShop on the
500 official test tasks with global indices 0--499. No evaluation instance is
used for supervised or reinforcement learning.

For each AIME~2026 problem and workflow condition, we sample eight completions
using temperature 0.7 and top-$p$ 0.95. We extract the final answer from each
completion and use majority voting for the problem-level prediction. Each
reported accuracy therefore aggregates 240 generations into 30 predictions.

For each WebShop task and workflow condition, we execute one complete
environment episode using greedy decoding ($T=0$ and top-$p=1$), a fixed
environment seed, and a budget of at most 14 model actions. Each reported
success rate is computed from 500 task-level trajectories. In both domains,
all workflow conditions use the same problem or task identifiers.

\section{Additional Results and Ablations}

\subsection{Condition-Level Results}
\label{app:detailed_results}

Tables~\ref{tab:webshop-all-workflow-regimes} and
\ref{tab:aime2026-all-workflow-regimes} expand the aggregated Partial and
Mixed scores in Table~\ref{tab:box2-main} into individual change points.
\textit{Good-$k$} removes guidance after the first $k$ good steps, whereas
\textit{$x$G+$y$B} replaces the remaining guidance with a bad suffix. This
breakdown shows how models respond as the useful prefix grows and the
misleading suffix shortens.

Following the evaluation protocol in
Appendix~\ref{app:evaluation-protocol}, all conditions reuse the same tasks,
environments, evaluators, and frozen workflows. Comparisons within each row
therefore isolate the effect of changing workflow availability and reliability.
The main-text scores $S_P$ and $S_M$ average the three Partial and Mixed
conditions reported here, respectively. These condition-level results also
determine the four paired effects $\Delta_{\mathrm{use}}$,
$\Delta_{\mathrm{bad}}$, $\Delta_{\mathrm{stop}}$, and
$\Delta_{\mathrm{switch}}$.

\begin{table*}[!htbp]
\centering
\caption{
WebShop accuracy across all workflow conditions.
\textit{Good-$k$} retains the first $k$ steps of the good workflow.
\textit{$x$G+$y$B} concatenates the first $x$ good steps with the
last $y$ bad steps.
}
\label{tab:webshop-all-workflow-regimes}

\small
\setlength{\tabcolsep}{5pt}
\renewcommand{\arraystretch}{1.08}

\begin{tabular}{lccccccccc}
\toprule

\multirow{2}{*}{\textbf{Model}}
&
\multicolumn{1}{c}{\textbf{Baseline}}
&
\multicolumn{2}{c}{\textbf{Full}}
&
\multicolumn{3}{c}{\textbf{Partial}}
&
\multicolumn{3}{c}{\textbf{Mix}}
\\

\cmidrule(lr){2-2}
\cmidrule(lr){3-4}
\cmidrule(lr){5-7}
\cmidrule(lr){8-10}

&
\cellcolor{baselinebg}\textit{None}
&
\textit{Good}
&
\textit{Bad}
&
\textit{Good-2}
&
\textit{Good-4}
&
\textit{Good-6}
&
\textit{2G+6B}
&
\textit{4G+4B}
&
\textit{6G+2B}
\\

\midrule

\rowcolor{benchmarkbg}
\multicolumn{10}{l}{
    \rule{0pt}{2ex}
    \small \# \textit{Search --- WebShop}
}\\

\textbf{Qwen3.5-4B}
    & \cellcolor{baselinebg} 20.8\%
    & 35.8\%
    & 15.2\%
    & 28.2\%
    & 35.8\%
    & 36.0\%
    & 27.4\%
    & 30.8\%
    & 32.4\%
\\

\textbf{Qwen3.5-9B}
    & \cellcolor{baselinebg} 25.2\%
    & 35.6\%
    & 15.4\%
    & 28.4\%
    & 34.8\%
    & 35.0\%
    & 27.8\%
    & 33.0\%
    & 37.0\%
\\

\textbf{Qwen3.5-27B}
    & \cellcolor{baselinebg} 28.6\%
    & 40.8\%
    & 17.2\%
    & 35.6\%
    & 40.2\%
    & 41.2\%
    & 38.2\%
    & 39.2\%
    & 41.0\%
\\

\bottomrule
\end{tabular}
\end{table*}

\begin{table*}[!htbp]
\centering
\caption{
AIME2026 accuracy across all workflow conditions.
\textit{Good-$k$} retains the first $k$ steps of the good workflow.
\textit{$x$G+$y$B} concatenates the first $x$ good steps with the
last $y$ bad steps.
}
\label{tab:aime2026-all-workflow-regimes}

\small
\setlength{\tabcolsep}{5pt}
\renewcommand{\arraystretch}{1.08}

\begin{tabular}{lccccccccc}
\toprule

\multirow{2}{*}{\textbf{Model}}
&
\multicolumn{1}{c}{\textbf{Baseline}}
&
\multicolumn{2}{c}{\textbf{Full}}
&
\multicolumn{3}{c}{\textbf{Partial}}
&
\multicolumn{3}{c}{\textbf{Mix}}
\\

\cmidrule(lr){2-2}
\cmidrule(lr){3-4}
\cmidrule(lr){5-7}
\cmidrule(lr){8-10}

&
\cellcolor{baselinebg}\textit{None}
&
\textit{Good}
&
\textit{Bad}
&
\textit{Good-2}
&
\textit{Good-4}
&
\textit{Good-6}
&
\textit{2G+6B}
&
\textit{4G+4B}
&
\textit{6G+2B}
\\

\midrule

\rowcolor{benchmarkbg}
\multicolumn{10}{l}{
    \rule{0pt}{2ex}
    \small \# \textit{Math --- AIME2026}
}\\

\textbf{Qwen3-0.6B}
    & \cellcolor{baselinebg} 16.7\%
    & 23.3\%
    & 20.0\%
    & 16.7\%
    & 16.7\%
    & 13.3\%
    & 13.3\%
    & 16.7\%
    & 16.7\%
\\

\textbf{Qwen3-4B}
    & \cellcolor{baselinebg} 66.7\%
    & 80.0\%
    & 46.7\%
    & 76.7\%
    & 86.7\%
    & 80.0\%
    & 53.3\%
    & 73.3\%
    & 73.3\%
\\

\textbf{Qwen3-8B}
    & \cellcolor{baselinebg} 76.7\%
    & 83.3\%
    & 50.0\%
    & 83.3\%
    & 83.3\%
    & 86.7\%
    & 70.0\%
    & 70.0\%
    & 83.3\%
\\

\bottomrule
\end{tabular}
\end{table*}

\subsection{Relative-Reward RL Ablation}
\label{app:relative-reward-rl}

The main experiments use $\mathrm{RL}_{\mathrm{env}}$ to optimize the
task-level outcome directly. We compare this objective with
$\mathrm{RL}_{\mathrm{rel}}$, which uses a relative reward.
Both variants start from the same SFT checkpoint and train only on
bad-workflow inputs, so the reward supplied to GRPO is the only difference.

For each workflow-conditioned input $x$, $\mathrm{RL}_{\mathrm{env}}$ samples
a rollout
$\tau \sim \pi_\theta(\cdot \mid x)$ and assigns it the terminal reward
$
r(\tau) = R(x,\tau),
$
where $R$ is the task evaluator. For mathematical reasoning, the reward is
binary answer correctness,
\begin{equation}
R_{\mathrm{math}}(x,\tau)
=
\mathbb{I}\!\left[\hat a(\tau)=a(x)\right],
\end{equation}
where $\hat a(\tau)$ is the extracted final answer. For WebShop, the reward is
the environment return,
\begin{equation}
R_{\mathrm{shop}}(x,\tau)
=
R_{\mathrm{env}}(\tau)\in[0,1].
\end{equation}
We retain tasks whose stochastic rollouts have non-degenerate returns and
optimize those groups with GRPO. The reward depends only on task success,
regardless of whether the rollout follows the supplied workflow.

\paragraph{Relative reward.}
The alternative objective measures improvement over the same model acting
without workflow guidance. At optimization step $t$, for each task $x$ and bad
workflow $W$, we sample $K=8$ workflow-conditioned trajectories and $K=8$
no-workflow reference trajectories from the same policy snapshot
$\pi_{\theta_t}$,
\begin{equation}
\tau^W_k \sim \pi_{\theta_t}(\cdot \mid x,W),
\qquad
\tau^0_j \sim \pi_{\theta_t}(\cdot \mid x),
\qquad k,j\in\{1,\ldots,K\}.
\end{equation}
Their mean task reward gives the contemporaneous no-workflow baseline
\begin{equation}
b_t(x)=\frac{1}{K}\sum_{j=1}^{K}r_{\mathrm{task}}(\tau^0_j),
\qquad
r_{\mathrm{rel}}(\tau^W_k)=r_{\mathrm{task}}(\tau^W_k)-b_t(x).
\end{equation}
The task reward is binary answer correctness for AIME~2026 and the native
environment reward in $[0,1]$ for WebShop. We sample both sets of trajectories
before the policy update and hold $b_t(x)$ fixed during that update. Policy
gradients come only from the workflow-conditioned trajectories. We use
$r_{\mathrm{rel}}$ directly as the policy advantage, without group-wise reward
normalization. Each workflow-conditioned rollout is therefore scored against
the mean no-workflow return from the same policy snapshot.

\begin{table*}[!htbp]
\centering
\caption{
WebShop success rates across workflow conditions for the outcome-based and
relative-reward RL variants.
\textit{Good-$k$} retains the first $k$ good steps, while
\textit{$x$G+$y$B} uses a length-$x$ good prefix followed by a length-$y$
bad suffix.
$\mathrm{RL}_{\mathrm{env}}$ uses the WebShop environment return, and
$\mathrm{RL}_{\mathrm{rel}}$ uses a same-snapshot relative reward.
}
\label{tab:webshop-training-stages-all-workflow-regimes}

\small
\setlength{\tabcolsep}{5pt}
\renewcommand{\arraystretch}{1.08}

\begin{tabular}{lccccccccc}
\toprule

\multirow{2}{*}{\textbf{Model}}
&
\multicolumn{1}{c}{\textbf{Baseline}}
&
\multicolumn{2}{c}{\textbf{Full}}
&
\multicolumn{3}{c}{\textbf{Partial}}
&
\multicolumn{3}{c}{\textbf{Mix}}
\\

\cmidrule(lr){2-2}
\cmidrule(lr){3-4}
\cmidrule(lr){5-7}
\cmidrule(lr){8-10}

&
\cellcolor{baselinebg}\textit{None}
&
\textit{Good}
&
\textit{Bad}
&
\textit{Good-2}
&
\textit{Good-4}
&
\textit{Good-6}
&
\textit{2G+6B}
&
\textit{4G+4B}
&
\textit{6G+2B}
\\

\midrule

\rowcolor{benchmarkbg}
\multicolumn{10}{l}{
    \rule{0pt}{2ex}
    \small \# \textit{Search --- WebShop}
}
\\

\textbf{Base}
    & \cellcolor{baselinebg} 25.2\%
    & 35.6\%
    & 15.4\%
    & 28.4\%
    & 34.8\%
    & 35.0\%
    & 27.8\%
    & 33.0\%
    & 37.0\%
\\

\textbf{+ SFT}
    & \cellcolor{baselinebg} 30.8\%
    & 33.6\%
    & 30.2\%
    & 32.2\%
    & 32.4\%
    & 33.4\%
    & 31.0\%
    & 32.2\%
    & 32.6\%
\\

\textbf{+ SFT + $\mathrm{RL}_{\mathrm{env}}$}
    & \cellcolor{baselinebg} 32.0\%
    & 34.8\%
    & 30.8\%
    & 32.8\%
    & 34.2\%
    & 34.2\%
    & 32.4\%
    & 32.8\%
    & 34.0\%
\\

\textbf{+ SFT + $\mathrm{RL}_{\mathrm{rel}}$}
    & \cellcolor{baselinebg} 29.6\%
    & 34.0\%
    & 31.4\%
    & 32.2\%
    & 33.6\%
    & 34.0\%
    & 32.0\%
    & 32.8\%
    & 34.0\%
\\

\bottomrule
\end{tabular}
\end{table*}

\begin{table*}[!htbp]
\centering
\caption{
AIME~2026 accuracy across workflow conditions for the outcome-based and
relative-reward RL variants.
Each entry reports majority-vote accuracy over eight samples per problem.
\textit{Good-$k$} retains the first $k$ good steps, while
\textit{$x$G+$y$B} uses a length-$x$ good prefix followed by a length-$y$
bad suffix.
$\mathrm{RL}_{\mathrm{env}}$ uses binary answer correctness, and
$\mathrm{RL}_{\mathrm{rel}}$ uses a same-snapshot relative reward.
}
\label{tab:aime2026-training-stages-all-workflow-regimes}

\small
\setlength{\tabcolsep}{5pt}
\renewcommand{\arraystretch}{1.08}

\begin{tabular}{lccccccccc}
\toprule

\multirow{2}{*}{\textbf{Model}}
&
\multicolumn{1}{c}{\textbf{Baseline}}
&
\multicolumn{2}{c}{\textbf{Full}}
&
\multicolumn{3}{c}{\textbf{Partial}}
&
\multicolumn{3}{c}{\textbf{Mix}}
\\

\cmidrule(lr){2-2}
\cmidrule(lr){3-4}
\cmidrule(lr){5-7}
\cmidrule(lr){8-10}

&
\cellcolor{baselinebg}\textit{None}
&
\textit{Good}
&
\textit{Bad}
&
\textit{Good-2}
&
\textit{Good-4}
&
\textit{Good-6}
&
\textit{2G+6B}
&
\textit{4G+4B}
&
\textit{6G+2B}
\\

\midrule

\rowcolor{benchmarkbg}
\multicolumn{10}{l}{
    \rule{0pt}{2ex}
    \small \# \textit{Mathematical Reasoning --- AIME 2026}
}
\\

\textbf{Base}
    & \cellcolor{baselinebg} 66.7\%
    & 80.0\%
    & 46.7\%
    & 76.7\%
    & 86.7\%
    & 80.0\%
    & 53.3\%
    & 73.3\%
    & 73.3\%
\\

\textbf{+ SFT}
    & \cellcolor{baselinebg} 73.3\%
    & 70.0\%
    & 66.7\%
    & 73.3\%
    & 76.7\%
    & 73.3\%
    & 63.3\%
    & 73.3\%
    & 73.3\%
\\

\textbf{+ SFT + $\mathrm{RL}_{\mathrm{env}}$}
    & \cellcolor{baselinebg} 73.3\%
    & 80.0\%
    & 56.7\%
    & 73.3\%
    & 66.7\%
    & 80.0\%
    & 70.0\%
    & 76.7\%
    & 76.7\%
\\

\textbf{+ SFT + $\mathrm{RL}_{\mathrm{rel}}$}
    & \cellcolor{baselinebg} 70.0\%
    & 80.0\%
    & 60.0\%
    & 76.7\%
    & 73.3\%
    & 83.3\%
    & 63.3\%
    & 66.7\%
    & 80.0\%
\\

\bottomrule
\end{tabular}
\end{table*}

\textbf{Results.}
Neither RL reward consistently dominates. Under both formulations, the trained
policies retain gains from good workflows and are less vulnerable to bad
workflows than the base model. On AIME~2026, $\mathrm{RL}_{\mathrm{rel}}$
improves utilization and robustness to workflows that are misleading from the
outset relative to $\mathrm{RL}_{\mathrm{env}}$, although it is more sensitive
when a useful prefix switches to a misleading suffix. On WebShop, relative
rewards yield slightly stronger paired workflow effects, while absolute
performance remains lower or comparable in most conditions. Because the two
variants share the same counterfactual SFT checkpoint and
bad-workflow-conditioned inputs, this comparison isolates the reward used for
RL. The results suggest that constructing the RL task around success despite
misleading guidance matters more than choosing between these two reward
formulations. Reward choice mainly changes the balance among utilization,
robustness, recovery, and absolute task performance. We use direct outcome
rewards in the main experiments because they are simple and competitive in
absolute performance, and report relative rewards as an objective ablation.

\newcommand{\BoxPrompt}[2]{%
  \begingroup
  \setlength{\fboxsep}{4.5pt}%
  \setlength{\fboxrule}{0.35pt}%
  \fbox{%
    \begin{minipage}[t]{\dimexpr\linewidth-2\fboxsep-2\fboxrule\relax}
      \small
      \setlength{\parindent}{0pt}%
      \setlength{\parskip}{0.35em}%
      \textbf{#1}\par
      \vspace{0.15em}
      #2
    \end{minipage}%
  }%
  \endgroup
}

\newcommand{\BoxField}[2]{%
  \textbf{#1} #2\par
}

\usetikzlibrary{arrows.meta,positioning}

\section{Harness Construction and Verification}
\label{app:harness-generation-prompts}

\subsection{Frontier Workflow Construction}
\label{app:workflow-construction}

\paragraph{Scope and model roles.}
We construct frontier workflows for OpenR1-Math, DeepSWE, BrowseComp,
and AutomationBench. The generator is
\texttt{qwen/qwen3.5-plus-20260420}. A separate model,
\texttt{openai/gpt-5.6-sol}, reviews the candidates with high reasoning
effort and aligns them with the construction requirements. We freeze the
step strings after this review has checked the intended distinction between
useful and misleading guidance.

\paragraph{Paired candidate generation.}
Generation has two phases. Phase~1 receives the public task and available
metadata and produces eight ordered, helpful steps,
$G=(G_1,\ldots,G_8)$. Phase~2 receives the same task and the fixed Good
candidate, then produces eight positionally aligned Bad steps,
$B=(B_1,\ldots,B_8)$. The Bad workflow must be a plausible, executable
procedure whose task-relevant errors are likely to compromise the result.
An alternative valid solution strategy does not qualify. Both phases use
only public task information; reference answers, hidden tests, seeded data,
and evaluator state are excluded. For mathematical tasks, the planner may
privately solve the public problem to check the proposed reasoning, but the
workflow cannot reveal the result.

\paragraph{Benchmark-specific Bad preparation.}
The frontier Bad templates follow a
\emph{preparation--error--propagation} structure. For OpenR1-Math,
$B_1$ is an accurate, low-information restatement. Steps $B_2$--$B_4$
may set up a plausible but inappropriate model, index convention,
conditioning choice, branch, or theorem frame, with the error introduced at
$B_5$. The setup must remain compatible with the public wording. It cannot
supply Good's correct intermediate derivation, assert a task-specific
recurrence or resolved branch set, or complete the erroneous calculation.

For DeepSWE, BrowseComp, and AutomationBench, $B_1$--$B_4$ provide
low-information preparation aligned with the stage of the task. These steps
cover generic inspection, baseline observation, organization of candidates or
records, and unresolved planning. They cannot copy or paraphrase Good's
task-specific localization, queries, discoveries, selected records, policy
mappings, or intermediate results. Each Bad step must also differ textually
from its aligned Good step.

Here \emph{weak} means that the setup provides little prescribed,
task-specific guidance. It does not require an independently false statement
at every setup position. Step $B_5$ must make a concrete erroneous decision or
transition. Steps $B_6$--$B_8$ carry the resulting state through execution,
inference, or locally consistent but misleading checks. A Bad workflow is
accepted only if it contains an identifiable task-relevant error and a
coherent path by which the guidance can compromise the result.

\paragraph{Coherent lead-in.}
A conspicuous contradiction or an easily repaired local error may let a
capable executor reject the instruction before engaging with the proposed
procedure. We therefore construct \emph{plausible procedural mistakes}. The
lead-in supplies an executable route to a consequential decision without
revealing Good's useful intermediate commitments. The error appears as
ordinary engineering, search, operational, or mathematical guidance and must
remain auditable from the public task. Whether an evaluated model detects and
overrides the error is measured during evaluation, not used as a construction
criterion.

\subsection{Semantic Review, Alignment, and Freezing}
\label{app:workflow-verification}

\paragraph{Review procedure.}
The reviewer examines the public task, the Good and Bad step strings, the
composed Mixed workflow, and the planner's failure-mode descriptions. It uses
the descriptions as audit annotations and independently checks that each
claimed error appears in the visible workflow and has the stated consequence
for the public task. When a candidate does not meet the construction
requirements, we revise it to align task fidelity, the error mechanism, and
execution dependencies before freezing. The review covers workflow goals and
specified transitions; it does not supervise every internal action of a target
model.

\paragraph{Acceptance criteria.}
\textbf{Good fidelity and usefulness.}
Good must preserve the public task constraints, use justified methods, and
provide feasible intermediate objectives and checks that materially help with
the task.

\textbf{Bad error and plausibility.}
Bad must contain an identifiable incorrect decision, inference, or state
transition with a task-relevant consequence. The full procedure must remain
natural, actionable, and internally coherent. Vague, omission-only, off-topic,
impossible, or self-announcing guidance is rejected.

\textbf{Prefix and continuation alignment.}
The early Bad preparation cannot reproduce Good's task-specific progress.
Later steps must preserve or propagate the erroneous commitment without
silently repairing it. The reviewer also checks that $B_5$--$B_8$ remain
executable from the state specified by $G_1$--$G_4$ and do not rely on state
available only from the Bad prefix.

\textbf{No leakage.}
Workflow steps cannot expose answers, hidden evaluation information,
treatment labels, or audit explanations. For mathematics, the proposed
misconception must change a necessary intermediate result, the admissible case
set, or the proof's validity. Errors that cancel out or leave the requested
result unchanged are rejected.

\paragraph{Task-specific error families.}
The templates use the error mechanisms listed below for construction and
semantic review. These are controlled benchmark failure modes; their inclusion
does not estimate how often the corresponding mistakes occur in human-written
workflows.

\begin{center}
\small
\renewcommand{\arraystretch}{1.15}
\begin{tabular}{@{}p{0.22\linewidth}@{\hspace{8pt}}p{0.70\linewidth}@{}}
\hline
\textbf{Domain} & \textbf{Error mechanisms} \\
\hline
OpenR1-Math &
Modeling, indexing, boundary, conditioning, branch, normalization,
theorem-applicability, and necessary-versus-sufficient errors; obvious
arithmetic slips are deliberately avoided. \\[0.35em]

DeepSWE &
Inappropriate subsystem or API choices; incorrect defaults or
operation ordering; compatibility and edge-case mistakes; misleading
test oracles and insufficient proxy validation. \\[0.35em]

BrowseComp &
Premature or incorrect entity binding; unsupported candidate filters;
relational, temporal, geographic, or quantifier mistakes; incorrect
joins and evidence-authority decisions. \\[0.35em]

AutomationBench &
Incorrect policy or source authority; eligibility, record, field,
recipient, or value mappings; inappropriate batching or side effects;
verification of an unrelated operational property. \\
\hline
\end{tabular}
\end{center}

\paragraph{Final workflows and solver visibility.}
Each domain contains 30 task-specific workflow bundles. We freeze the accepted
step strings and use the same strings for every evaluated model. Good and Bad
use all eight steps, Partial uses $G_{1:4}$, and Mixed uses
$(G_{1:4},B_{5:8})$. The evaluated solver sees only the selected
natural-language steps. Generation prompts, Good/Bad field names, failure-mode
annotations, reviewer explanations, and the instruction to construct
misleading guidance remain hidden from the solver.

The templates below summarize the generation instructions and shared review
criteria. They are not verbatim API messages.

\paragraph{Scope of verification.}
Semantic approval establishes that a workflow satisfies the construction
criteria. It does not guarantee a particular empirical outcome for a target
model. Because the review is model-based adjudication, we report neither an
inter-annotator reliability coefficient nor a candidate-level rejection rate.
The resulting benchmark evaluates a controlled collection of plausible
procedural errors and does not estimate their prevalence in human-authored
workflows.

\subsection{Semantic Reviewer Protocol and Audit Flow}
\label{app:semantic-reviewer-protocol}

\paragraph{Reviewer implementation.}
We separate generation from validation with a fixed semantic reviewer,
\texttt{openai/gpt-5.6-sol}, configured with high reasoning effort. This model
does not generate candidates. It receives the public task, the Good and Bad
workflows shown to the target model, the composed Mixed workflow, and the
planner's failure-mode descriptions. The reviewer treats those descriptions as
audit aids and verifies the claimed defect and its consequence against the
visible workflow and public task.

\begin{figure}
    \centering
    \includegraphics[width=1\linewidth]{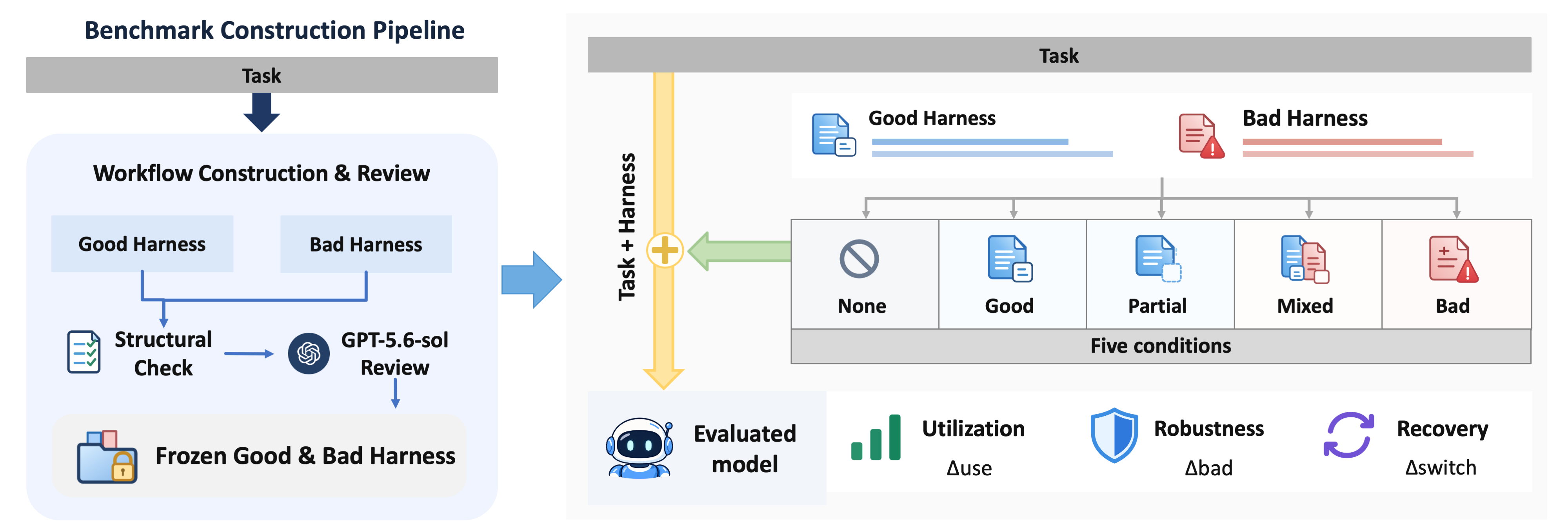}
    \caption{\textbf{Benchmark construction and evaluation pipeline.} For each task, we construct good and bad harnesses, verify their structure, review their content with GPT-5.6-sol, and freeze the resulting harnesses before evaluation. We evaluate the model with no harness or a good, partial, mixed, or bad harness to measure utilization ($\Delta_{\mathrm{use}}$), robustness ($\Delta_{\mathrm{bad}}$), and recovery after a reliability switch ($\Delta_{\mathrm{switch}}$).}
    \label{fig:benchmark-construction-evaluation}
\end{figure}

\paragraph{Benchmark construction pipeline.}
Construction proceeds through four stages: workflow generation, structural
validation, semantic audit and alignment, and freezing the workflow artifacts
for evaluation. Figure~\ref{fig:benchmark-construction-evaluation} summarizes
this construction process and the five conditions used to evaluate target-model
behavior under the frozen workflows.

\paragraph{Bundle-level audit.}
The reviewer applies the acceptance criteria to the complete bundle. Good must
contain feasible, useful, and task-relevant steps without leaking answers,
hidden evaluation information, or unsupported assumptions. Bad is assessed as
a complete workflow: every step need not be independently harmful, but the
chain must be task-relevant, executable, coherent, and likely to compromise the
result. Low-information early steps are allowed when the chain contains an
identifiable erroneous transition and a coherent causal path. For Mixed, the
reviewer checks that $B_5$--$B_8$ remain executable after $G_{1:4}$ and do not
depend on state available only from the Bad prefix.

\paragraph{Reviewer scope and limitations.}
The reviewer performs a semantic audit of the construction criteria. We do not
use a blind human annotation protocol. Empirical evaluation of target-model
behavior remains separate, and the audit does not estimate naturally occurring
human error distributions. We will release the complete frozen reviewer prompts and
schemas with the benchmark artifacts; this appendix gives an abridged account
of the audit protocol.

\section{Memory}

\subsection{Memory Extension Details}
\label{app:memory}

\paragraph{Evaluation setup.}
We evaluate transfer on the text-only subset of LongMemEval-V2-Small.
All checkpoints are Qwen3.5-9B variants obtained from our WebShop
training pipeline.
The SFT checkpoint is trained with WebShop bad-workflow supervision.
Starting from SFT, $RL_{\mathrm{env}}$ is further optimized using
WebShop environment rewards.
No LongMemEval-V2 data are used during either training stage.

We select 30 questions from 387 eligible text-only questions using
stratified sampling over domain and question type.
The subset contains 15 Web and 15 Enterprise questions and covers four
text-only memory-ability groups: 9 static-state recall, 7 dynamic-state
tracking, 5 workflow-knowledge, and 9 premise-awareness questions.
Environment-gotcha questions are excluded because the corresponding
items in the frozen benchmark are image-dependent.
Each model--condition pair is evaluated with three stochastic rollouts.

For every question, the Base, GOOD, and BAD conditions share the same
question, underlying archive, system prompt, retrieval interface, and
generation configuration.
Only the supplied memory brief differs.
Base provides no factual brief, GOOD provides an answer-sufficient
claim supported by the archive, and BAD contains one answer-bearing
atomic corruption while preserving the same underlying task and
archive.

\paragraph{Memory construction and verification.}
GOOD/BAD memory pairs are constructed and frozen before any target-model
evaluation.
Candidate pairs are generated from the official question, reference
answer, and the corresponding LongMemEval-V2-Small archive.
For 26 of the 30 questions, the candidate pair is generated with
DeepSeek-V4-Flash-0731 using read-only access to the archive; the
remaining four pairs are constructed manually with archive-grounded
evidence when automatic generation does not satisfy the construction
constraints.

Each pair is subsequently checked by deterministic schema and evidence
validation and independently verified with GLM-5.2 at temperature
zero.
The verifier receives the question, reference answer, candidate pair,
and relevant evidence from the official archive.
A pair is retained only when the GOOD memory is correct and
answer-sufficient, the BAD memory induces a concrete incorrect answer
through a single answer-bearing atomic corruption, the shared context
remains valid, and the official evidence supports GOOD while
contradicting BAD.
Treatment-revealing labels or instructions are excluded from all
memory briefs.
All accepted pairs are frozen before reader evaluation.

The resulting BAD memories contain 11 premise insertions, 9 value
substitutions, 8 entity substitutions, and 2 relation reversals.
GOOD and BAD memories are approximately length-matched, with mean
lengths of 54.97 and 52.53 words, respectively.

\paragraph{Tool-based inference.}
The evaluated checkpoint may either answer directly from the supplied
memory or independently verify it against the same read-only archive.
We provide a bounded retrieval interface that allows the model to
search the archive and inspect relevant content as needed.
Retrieved information is appended to the same conversation, after
which the model may continue retrieval or produce its final answer.
Thus, retrieval decisions and final answering are performed by the
same evaluated Qwen3.5-9B checkpoint rather than by separate controller
and reader models.

We allow at most eight archive retrieval calls and ten interaction
turns.
Generation uses temperature $0.6$, top-$p$ $0.95$, and top-$k$ $20$,
with thinking enabled.
The controller-generation budget is 8,192 tokens and the forced
final-answer budget is 20,000 tokens.
We use rollout seeds 303, 304, and 305.
The service supports a maximum model context of 262,144 tokens.
The 200K memory-context limit from LongMemEval-V2 is retained as a
capacity constraint; unlike the standard context-gathering protocol,
however, the archive remains external and is accessed incrementally
through the bounded read-only retrieval interface.

\paragraph{Evaluation and reported metrics.}
Score in Table~\ref{tab:memory-transfer} denotes answer accuracy.
We follow the released LongMemEval-V2 scoring implementation, including
its answer extraction and \textsc{unknown} handling.
Of the 30 selected questions, 16 are evaluated by normalized
phrase matching, 5 by multiple-choice matching, and 9 by abstention
evaluation.
For the abstention items, we use the released LongMemEval-V2 judge
protocol with GLM-5.2 as the judge model.

Tool Rate denotes the fraction of trajectories containing at least one
archive retrieval operation.
All checkpoints access the archive in all Base trajectories, as no
factual memory brief is supplied in that condition.
The GOOD/BAD comparison therefore measures whether the model changes
its verification behavior according to the reliability of the supplied
memory.
Because our setting introduces controlled memory interventions and
interactive archive access, these numbers are intended as a transfer
evaluation built on LongMemEval-V2-Small and are not directly
comparable to the official leaderboard protocol.

\paragraph{Additional retrieval analysis.}
The difference between GOOD and BAD is also reflected in retrieval
intensity.
For $RL_{\mathrm{env}}$, the mean number of archive retrieval calls
increases from 3.62 under GOOD memory to 4.40 under BAD memory, compared
with 3.63 versus 3.88 for Base and 4.38 versus 4.63 after SFT.
The main paper reports only Tool Rate for simplicity.

Archive retrieval is also associated with successful correction under
corrupted memory.
Under BAD memory, every correct answer across all three checkpoints is
produced by a trajectory that accesses the archive:
9/9 correct Base trajectories, 13/13 correct SFT trajectories, and
12/12 correct $RL_{\mathrm{env}}$ trajectories use archive retrieval.
Moreover, all of these successful trajectories retrieve evidence
consistent with the frozen supporting evidence for the corresponding
question.
No BAD trajectory that avoids archive retrieval produces a correct
answer.
This pattern supports interpreting increased retrieval under BAD memory
as verification behavior rather than merely additional tool activity.

\section{Case}

\subsection{Case Studies Across Contexts}
\label{app:case_studies_contexts}

We present three representative trajectories to illustrate a common
behavioral shift after training.
Across single-model workflow execution, multi-agent collaboration, and
memory-augmented reasoning, the trained model is less likely to preserve
fallible context and more likely to re-check, verify, or override it.
These cases are qualitative illustrations rather than exhaustive evidence;
the corresponding aggregate trends are reported in the main text.

\begin{figure}[h]
    \centering
    \includegraphics[width=0.74\linewidth]{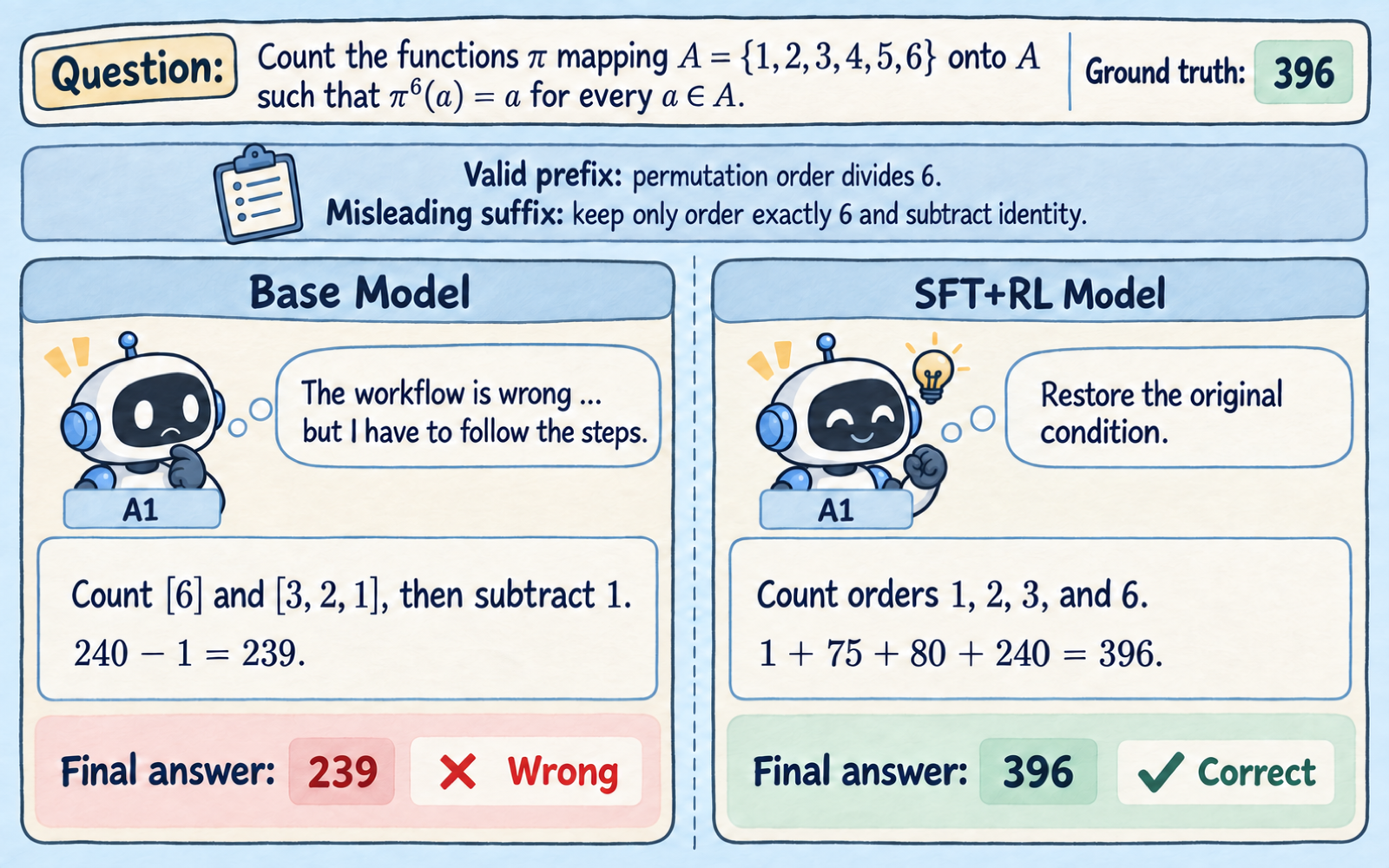}
    \caption{
    \textbf{Single-model workflow case.}
    The task is to count permutations of $\{1,\ldots,6\}$ whose order divides~6.
    The shared mixed workflow contains a valid prefix but a misleading suffix
    that replaces ``divides 6'' with ``equals 6''.
    The Base model follows the misleading suffix and outputs an incorrect answer,
    whereas the trained model restores the original condition and recomputes the
    correct result.
    Responses shown in the figure are abridged for visualization.
    }
    \label{fig:case_single}
\end{figure}

In this case (Figure~\ref{fig:case_single}), the shared workflow first identifies the correct divisibility condition and then silently switches to the stronger and incorrect requirement that the permutation order be exactly 6. The Base trajectory recognizes the inconsistency but still follows the misleading suffix, whereas the trained model restores the original condition and counts all valid permutations. In a separate diagnostic on the same problem and the Mixed workflow, we draw 16 samples from each checkpoint. Base solves 1/16 samples correctly, SFT solves 8/16, and SFT+$RL_{\mathrm{env}}$ solves 16/16. These per-sample diagnostic counts are separate from the main AIME evaluation, which uses eight samples per problem--condition pair and reports majority-vote accuracy.

\begin{figure}[h]
    \centering
    \includegraphics[width=0.74\linewidth]{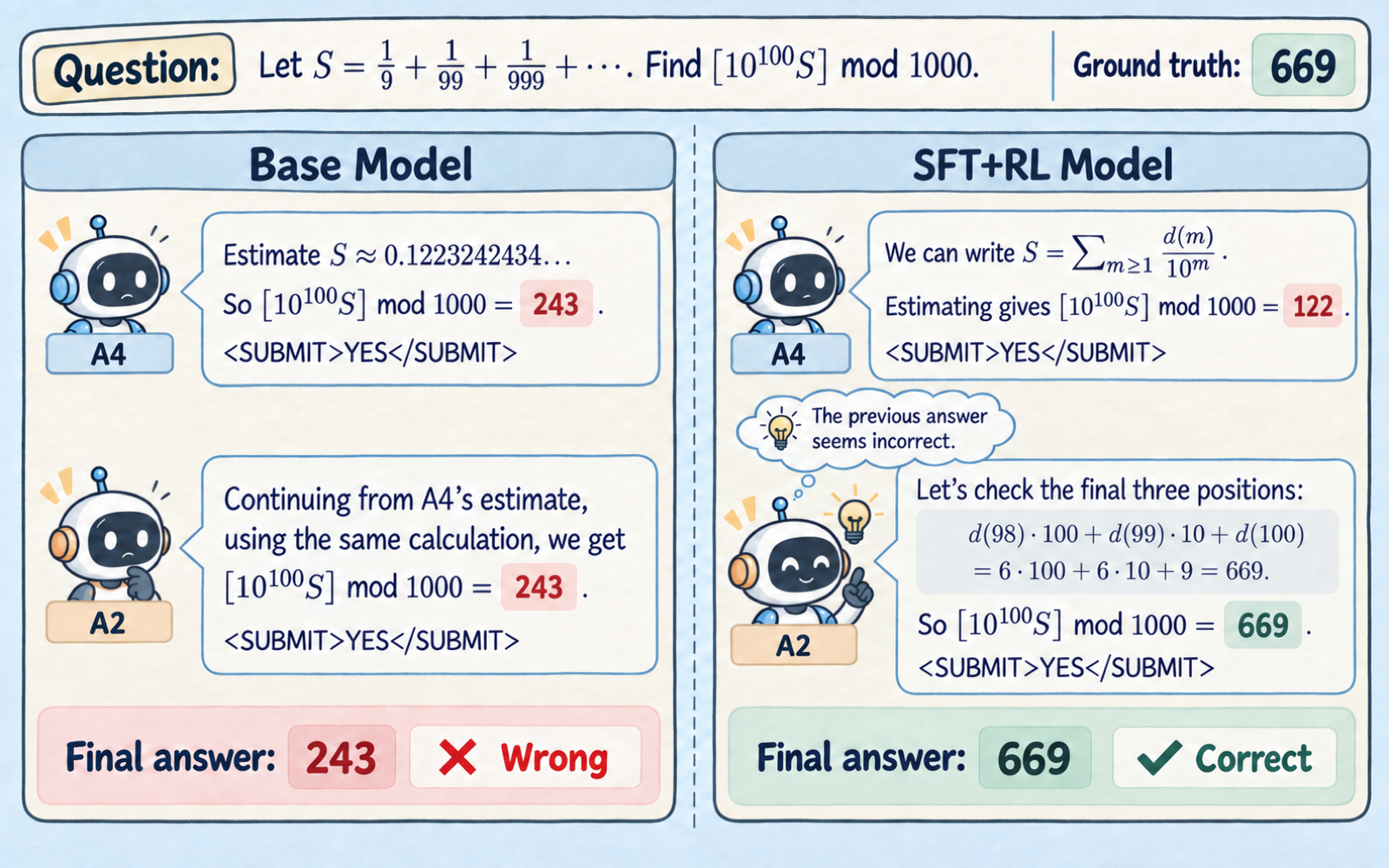}
    \caption{
    \textbf{Multi-agent collaboration case.}
    We compare Base and trained checkpoints on the same AIME~2026 problem
    under the same A4$\rightarrow$A2 handoff pattern.
    In both trajectories, the first agent produces an incorrect public draft.
    The Base second agent continues from the inherited estimate and preserves
    the wrong answer, whereas the trained second agent re-examines the draft
    and corrects the answer.
    Text in the figure is abridged; the thought bubble is a schematic
    visualization rather than a verbatim model output.
    }
    \label{fig:case_eom}
\end{figure}

This trajectory (Figure~\ref{fig:case_eom}) illustrates cross-agent error correction.
The first agent is wrong in both conditions, so the comparison isolates how
the second agent responds to inherited peer context.
Base preserves the earlier conclusion, while the trained model recomputes the
decisive terms and corrects the final answer to 669.
The qualitative pattern is consistent with our aggregate recovery analysis,
where trained checkpoints are less likely than Base to preserve an initial
wrong answer from a previous agent.

\newpage
\begin{figure}[h]
    \centering
    \includegraphics[width=0.74\linewidth]{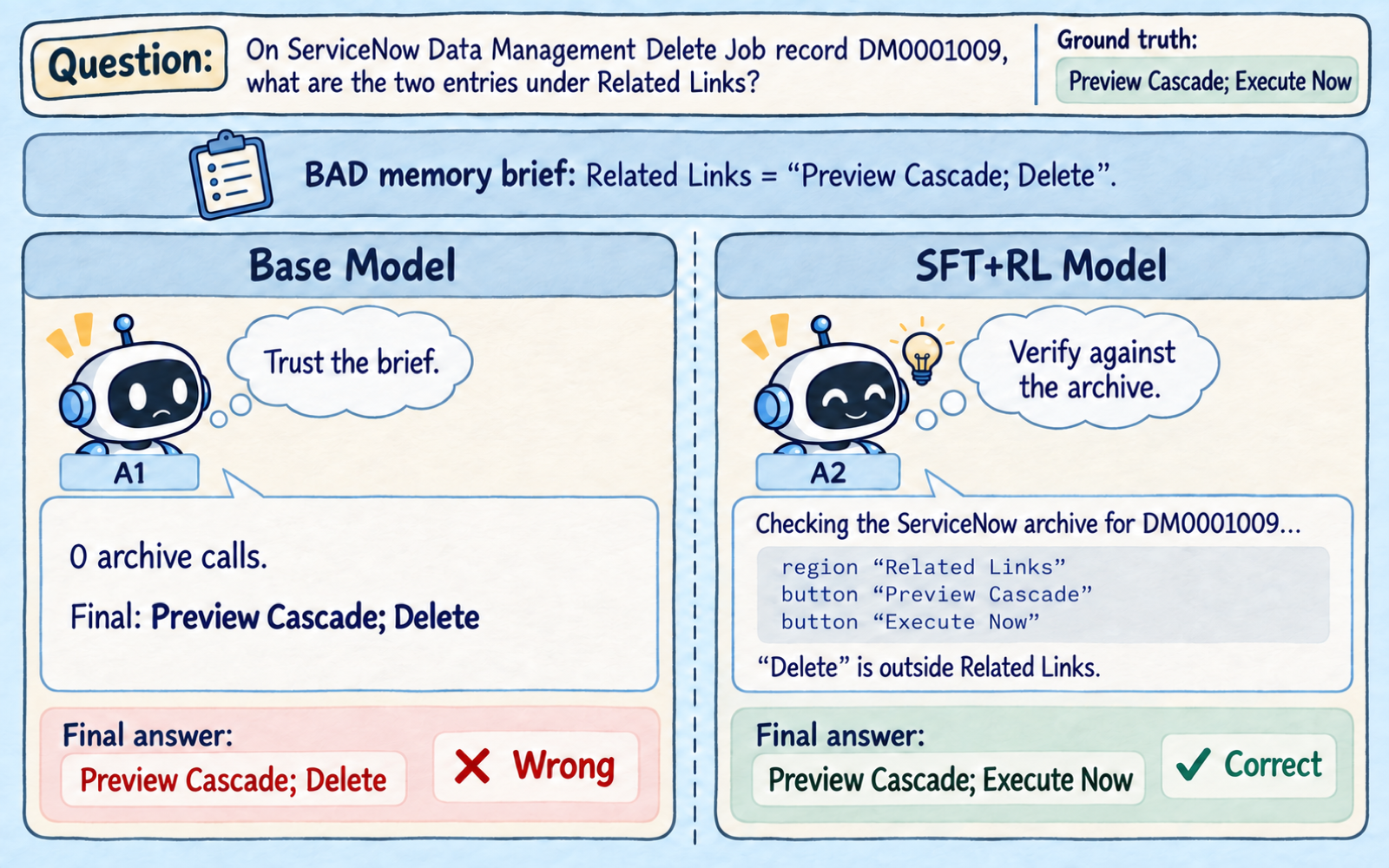}
    \caption{
    \textbf{Memory-augmented reasoning case.}
    Both checkpoints receive the same corrupted memory brief for the same
    ServiceNow question.
    The Base model copies the corrupted memory directly, while the trained
    model verifies the claim against archive evidence and replaces the wrong
    item with the correct one.
    The figure shows an abridged view of the interaction; detailed task
    description is provided in the text.
    }
    \label{fig:case_memory}
\end{figure}

In this example (Figure~\ref{fig:case_memory}), the BAD memory brief replaces one of the true entries under
\texttt{Related Links} with a plausible but incorrect control.
Base directly trusts the brief and returns the corrupted item.
The trained model instead queries the archive, observes that the incorrect
item lies outside the relevant region, and answers correctly.
A GOOD-memory control confirms that the trained behavior is not blanket
rejection: when the brief is reliable, both checkpoints answer correctly
without additional verification.

Across all three cases, the gain is not merely higher answer accuracy.
The common shift is behavioral: the trained model more often checks
inherited or retrieved context against task constraints and available
evidence, instead of directly propagating a misleading signal.
These case studies therefore complement the main quantitative results by
making the mechanism of selective reliance more explicit.

\clearpage
\section{Abridged Generation Prompts}
\label{app:workflow-prompt-templates}

\paragraph{Common output contract.}
Phase~1 returns an eight-element \texttt{steps} array.
Phase~2 returns eight \texttt{pairs}, each containing
\texttt{index}, \texttt{good\_step}, \texttt{bad\_step}, and
\texttt{failure\_mode}.
The Good field copies the fixed candidate exactly, and every Bad step must
differ from its aligned Good counterpart. Failure-mode descriptions are used
only during review. The panels distinguish useful progress in Good from the
preparation--error--propagation structure used in Bad.

\subsubsection*{OpenR1-Math: Mathematical Workflows}
\label{app:prompt-openr1-math}

\noindent
\textbf{Inputs.}
Phase~1 receives the public mathematical problem and metadata. Phase~2 also
receives the frozen Good workflow $G$.
Neither a correct nor an intentionally incorrect concrete final answer
may appear in the visible workflow.

\medskip

\noindent
\begin{minipage}[t]{0.485\linewidth}
\vspace{0pt}
\BoxPrompt{Phase 1---Good workflow}{
\BoxField{Objective.}{
Generate eight helpful reasoning steps.
Privately derive a correct solution from the public problem to check the
plan without exposing the resulting answer.}

\BoxField{Mathematical fidelity.}{
Preserve all values, signs, domains, quantifiers, definitions,
relations, and requested outputs.
Do not add unstated positivity, integrality, uniqueness, regularity, or
endpoint assumptions.}

\BoxField{Useful setup: $G_1$--$G_4$.}{
Formalize the givens and target; select necessary justified machinery;
derive a non-decisive symbolic relation; and record unresolved cases,
domains, or an unevaluated expression ready for resolution.
Preserve every feasible branch.}

\BoxField{Completion.}{
Use $G_5$ for the first decisive computation or comparison, $G_6$ for
cases and restrictions, $G_7$ for independent verification, and $G_8$
for formatting the solver's own result.
Adapt these stages to proof tasks when appropriate.}

\BoxField{Exact reasoning.}{
Prefer a minimal justified method and symbolic equivalence.
Track signs, denominator restrictions, boundaries, orientations,
units, equality cases, and extraneous solutions.}

\BoxField{No answer leakage.}{
Do not provide a final value, selected option, completed recurrence or
case set, decisive arithmetic, or final inequality.
State symbolic objectives and leave the derivation to the solver.}
}
\end{minipage}
\hfill
\begin{minipage}[t]{0.485\linewidth}
\vspace{0pt}
\BoxPrompt{Phase 2---Bad workflow}{
\BoxField{Objective and private check.}{
Deliberately construct eight plausible instructions likely to produce a
wrong result or invalid proof.
Privately derive the correct solution and reject a mutation that is
actually correct, cancels later, or leaves the requested result
unchanged.}

\BoxField{Weak/preparatory Bad setup: $B_1$--$B_4$.}{
Make $B_1$ an accurate low-information restatement.
$B_2$--$B_4$ may prepare a plausible but inappropriate modeling,
indexing, conditioning, branch, or theorem frame.
Do not copy or paraphrase Good's correct equations, bounds, branch
reductions, or intermediate results.
Keep the preparation compatible with the public wording; do not assert
a task-specific recurrence, sign convention, resolved branch set, or
bound.
Avoid overt contradictions and elementary early slips.}

\BoxField{Decisive misconception: $B_5$.}{
Apply one concrete structural error to the symbolic state left by
$G_4$.
Prefer a modeling, indexing, boundary, conditioning, branch,
normalization, or theorem-hypothesis error that requires revisiting the
setup to diagnose.
Do not simply contradict a preceding equation or reverse a stated
value.}

\BoxField{Propagation: $B_6$--$B_8$.}{
Reuse the resulting symbolic state in later steps and verify within
that interpretation.
Do not add an unrelated error, restart from the original givens, or
silently restore the correct interpretation.
The suffix must remain usable after the Good prefix.}

\BoxField{Review requirements.}{
Reject obvious arithmetic slips, malformed expansions, false one-line
identities, and self-announcing instructions.
Keep results symbolic and place the explicit failure explanation only
in audit annotations.}
}
\end{minipage}

\clearpage

\subsubsection*{DeepSWE: Repository Workflows}
\label{app:prompt-deepswe}

\noindent
\textbf{Inputs.}
Phase~1 receives the public repository task and metadata. Phase~2 also receives
the frozen Good workflow $G$.
Only generator-facing instructions are shown below.

\medskip

\noindent
\begin{minipage}[t]{0.485\linewidth}
\vspace{0pt}
\BoxPrompt{Phase 1---Good workflow}{
\BoxField{Objective.}{
Generate eight ordered and helpful repository-level execution steps
from public issue information.}

\BoxField{Task fidelity.}{
Preserve public APIs, defaults, ordering constraints, compatibility,
exports, tests, examples, and delivery requirements.
Do not invent files, symbols, dependencies, commands, or acceptance
criteria before inspection establishes them.}

\BoxField{Useful setup: $G_1$--$G_4$.}{
Provide concrete reconnaissance, localization, design, or
implementation objectives appropriate to the task.
Leave a coherent repository state from which an autonomous coding agent
can continue.
Do not claim a particular file or test already exists without
inspection.}

\BoxField{Completion and checks.}{
Cover implementation, integration, edge cases, compatibility, focused
tests, broader validation, and final delivery checks.
Adapt the stages rather than forcing unnecessary work.}

\BoxField{Execution freedom.}{
State intermediate objectives and observable checks.
Prefer examining existing code and tests before edits, and validate
proportionally to the change.
Leave commands and implementation choices to the coding agent.}

\BoxField{No solution leakage.}{
Do not provide patches, exact replacement code, reference solutions,
hidden tests, or evaluator-specific assertions.
Remove unsupported assumptions and premature implementation
commitments.}
}
\end{minipage}
\hfill
\begin{minipage}[t]{0.485\linewidth}
\vspace{0pt}
\BoxPrompt{Phase 2---Bad workflow}{
\BoxField{Objective.}{
Deliberately construct eight aligned instructions whose complete
procedure is plausible and executable but likely to lead to a
task-relevant implementation failure if followed.}

\BoxField{Weak Bad setup: $B_1$--$B_4$.}{
Use generic reconnaissance, baseline observation, and unresolved edit
planning.
Do not copy or paraphrase Good's task-specific paths, symbols, APIs,
tests, localization, design decisions, or intermediate findings.
Keep this preparation feasible and textually distinct from Good.}

\BoxField{Error-bearing commitment: $B_5$.}{
Specify a concrete implementation or state-transition error that can
operate on the state left by $G_4$.
Prefer a plausible compatibility, default, ordering, or edge-case
mistake rather than an overt instruction to violate the issue.}

\BoxField{Propagation: $B_6$--$B_8$.}{
Carry the error into testing, validation, and finalization.
Use a mistaken local test oracle and a focused proxy that accepts the
affected behavior.
At least two later positions should actively preserve or validate the
error rather than conduct a broad specification audit that restores the
correct implementation.}

\BoxField{Review requirements.}{
The complete Bad chain and the continuation after $G_{1:4}$ must remain
coherent and likely harmful.
Do not rely on incompatible Bad-only state.
Express the mistake as ordinary engineering guidance and keep treatment
labels and failure explanations outside the visible steps.}
}
\end{minipage}

\clearpage

\subsubsection*{BrowseComp: Information-Search Workflows}
\label{app:prompt-browsecomp}

\noindent
\textbf{Inputs.}
Phase~1 receives the public question. Phase~2 also receives the frozen Good
workflow $G$. Both phases use only the content of the public question.

\medskip

\noindent
\begin{minipage}[t]{0.485\linewidth}
\vspace{0pt}
\BoxPrompt{Phase 1---Good workflow}{
\BoxField{Objective.}{
Generate eight helpful information-search steps while preserving the
public question exactly.}

\BoxField{Constraint fidelity.}{
Retain every entity role, modifier, quantifier, date phrase, range,
relation, geographic constraint, and requested output.
Do not invent numeric endpoints for fuzzy dates, change a stated
relation into a superficially similar one, or impose an unsupported
nearest-entity condition.}

\BoxField{Useful setup: $G_1$--$G_4$.}{
Extract constraints, decompose queries, discover candidate sets, and
capture source provenance.
Leave a traceable candidate-and-evidence state for continued search and
retain alternatives until evidence resolves them.}

\BoxField{Resolution and verification.}{
Resolve entities across sources, explicitly check relations, dates,
and quantifiers, and corroborate ambiguous joins.
Address conflicts before formatting the answer.}

\BoxField{Evidence standards.}{
Prefer primary or authoritative sources for decisive facts.
Search snippets may support discovery but not a final evidential join.
Source preferences must not become invented task restrictions.}

\BoxField{No answer leakage.}{
Do not guess the final answer or claim that a source has been found
before searching.
Remove unsupported assumptions and premature candidate commitments.}
}
\end{minipage}
\hfill
\begin{minipage}[t]{0.485\linewidth}
\vspace{0pt}
\BoxPrompt{Phase 2---Bad workflow}{
\BoxField{Objective.}{
Deliberately construct eight plausible search instructions whose
complete procedure is likely to yield a wrong candidate, unsupported
relation, or unjustified answer when followed.}

\BoxField{Weak Bad setup: $B_1$--$B_4$.}{
Use generic source selection, candidate tracking, evidence organization,
and unresolved planning.
Do not copy Good's task-specific queries, candidates, identifiers,
disambiguating facts, joins, discoveries, or intermediate conclusions.
Do not introduce a wrong constraint or provide a task-specific
information shortcut beyond solving from the public question alone.}

\BoxField{Error-bearing binding: $B_5$.}{
Introduce an auditable but locally plausible error in entity binding,
source authority, join direction, boundary convention, geography, or
evidence resolution.
It must be directly applicable to the candidate/evidence state left by
$G_4$.}

\BoxField{Propagation: $B_6$--$B_8$.}{
Search and corroborate conditional on the mistaken binding.
At least two later steps must preserve, amplify, or incorrectly
validate it.
Do not reopen all candidates or rerun a full constraint audit that
repairs the misconception.}

\BoxField{Review requirements.}{
Both the complete Bad procedure and the continuation after $G_{1:4}$
must remain coherent and likely harmful.
Avoid conspicuous negations, answer leakage, and
treatment-revealing wording.
An audit-only failure claim must correspond to a defect present in the
visible instructions.}
}
\end{minipage}

\clearpage

\subsubsection*{AutomationBench: Tool-Execution Workflows}
\label{app:prompt-automationbench}

\noindent
\textbf{Inputs.}
Phase~1 receives the public system or user task, metadata, and listed tools.
Phase~2 also receives the frozen Good workflow $G$.

\medskip

\noindent
\begin{minipage}[t]{0.485\linewidth}
\vspace{0pt}
\BoxPrompt{Phase 1---Good workflow}{
\BoxField{Objective.}{
Generate eight helpful tool-execution steps grounded in the public task
and available tools.}

\BoxField{Task fidelity.}{
Preserve scope, policy criteria, record fields, exact values,
recipients, destinations, ordering, notifications, and required side
effects.
Do not invent thresholds, records, resources, or tool capabilities.}

\BoxField{Preparation before action.}{
Use authoritative-source discovery, schema inspection, independent
per-record reasoning, and a traceable no-write action preview.
Retain stable record identifiers and exact source values before the
corresponding writes, and do not pretend that a lookup has already
occurred.}

\BoxField{Correct execution.}{
Use read-before-write behavior and narrowly scoped updates.
Apply policy criteria to every in-scope record, preserve ambiguous
cases instead of guessing, and complete required notifications and
downstream records.}

\BoxField{Grounding and verification.}{
Use resource names supplied by the public task or refer to retrieved
objects by role after schema inspection.
Re-read changed records and check completeness, required side effects,
and idempotency.}

\BoxField{No leakage or invented actions.}{
Do not use hidden assertions, seeded data, evaluator state, or an
answer key.
Respect communication rules, exact-value requirements, and the
requested action scope.}
}
\end{minipage}
\hfill
\begin{minipage}[t]{0.485\linewidth}
\vspace{0pt}
\BoxPrompt{Phase 2---Bad workflow}{
\BoxField{Objective.}{
Deliberately construct eight executable instructions whose complete
procedure is likely to produce incorrect records, communications,
values, or operational state when followed.}

\BoxField{Weak Bad setup: $B_1$--$B_4$.}{
Use generic discovery, read-before-write reasoning, per-record
consideration, and an unresolved preview.
Do not copy Good's selected records, identifiers, recipients,
destinations, exact values, mappings, or retrieved policy facts.
Do not introduce unauthorized actions or task-specific shortcuts
relative to solving from the public task.}

\BoxField{Error-bearing action: $B_5$.}{
Make a concrete, plausible error in a field, recipient, join, batch
rule, default, or policy-to-action mapping.
The action must operate on the plan/state specified by $G_4$ and affect
a requested task outcome.}

\BoxField{Propagation: $B_6$--$B_8$.}{
Carry the affected state into required side effects and then verify
proximate fields or counts that remain compatible with it.
At least two later steps must preserve, propagate, or falsely validate
the error.
Do not rerun the original policy audit in a way that restores the
correct state.}

\BoxField{Review requirements.}{
Use actually retrieved objects and fields.
When a literal name is not public, use a deterministic rule over the
returned schema rather than inventing a resource.
The suffix must not depend on records selected only by the Bad prefix.
Keep the visible wording natural and exclude treatment labels and audit
explanations.}
}
\end{minipage}

\end{document}